\documentclass{tlp}

\usepackage{float}
\usepackage{graphicx}
\usepackage{caption}
\usepackage{subcaption}
\usepackage{multirow}

\usepackage{amsmath}  
\usepackage{verbatim} 
\usepackage{enumerate}
\usepackage{amsfonts}
\usepackage{amssymb}
\usepackage{graphicx}
\usepackage{booktabs}
\usepackage{paralist}
\usepackage{url}
\newcommand{\urlBiBTeX}[1]{\url{#1}}

\DeclareMathOperator{\val}{=}  
\DeclareMathOperator{\notval}{\neq}  

\def\happens{\textsf{\small happensAt}}

\def\initially{\textsf{\small initially}}
\def\holdsAt{\textsf{\small holdsAt}}

\def\initiatedAt{\textsf{\small initiatedAt}}
\def\terminatedAt{\textsf{\small terminatedAt}}

\def\broken{\textsf{\small broken}}
\def\negate{\textsf{\small negate}}

\def\mike{\textsf{\small mike}}
\def\sarah{\textsf{\small sarah}}

\def\nbf{\textsf{\small not}}
\def\pnot{\textsf{\small problog\_not}}
\def\pneg{\textsf{\small problog\_neg}}
\def\true{\textsf{\small true}}
\def\false{\textsf{\small false}}

\newenvironment{mysplit}%
  {\arraycolsep 0pt \begin{array}{l}}%
  {\end{array}}

\def\walking{$\mathit{walking}$}
\def\activeb{$\mathit{active}$}
\def\inactive{$\mathit{inactive}$}
\def\running{$\mathit{running}$}
\def\abrupt{$\mathit{abrupt}$}

\def\meet{$\mathit{meeting}$}
\def\fight{$\mathit{fighting}$}
\def\leave{$\mathit{leaving\_object}$}
\def\move{$\mathit{moving}$}

\def\person{$\mathit{person}$} 
\def\close{$\mathit{close}$}

\title[A Probabilistic Logic Programming Event Calculus]{A Probabilistic Logic Programming \\  Event Calculus}

\author[A. Skarlatidis et al.]{Anastasios Skarlatidis$^{1,2}$, Alexander Artikis$^1$, Jason Filippou$^{1,3}$ and Georgios Paliouras$^1$ \\ 
$^1$Institute of Informatics and Telecommunications, NCSR Demokritos, Athens, Greece \\
$^2$Department of Digital Systems, University of Piraeus, Greece \\
$^3$University of Maryland, USA \\
\email{\{anskarl, a.artikis, jfilip, paliourg\}@iit.demokritos.gr }}

 \submitted{06 April 2012}
 \revised{10 November 2012}
 \accepted{07 March 2013} 

\begin{document}
\maketitle



\noindent\textbf{Note:} The paper has been accepted for publication in the Theory and Practice of Logic Programming (TPLP) Journal.

\begin{abstract}
We present a system for recognising human activity given a symbolic representation of video content. The input of our system is a set of time-stamped short-term activities (STA) detected on video frames.  The output is a set of recognised long-term activities (LTA), which are pre-defined temporal combinations of STA. The constraints on the STA that, if satisfied, lead to the recognition of a LTA, have been expressed using a dialect of the Event Calculus. In order to handle the uncertainty that naturally occurs in human activity recognition, we adapted this dialect to a state-of-the-art probabilistic logic programming framework. We present a detailed evaluation and comparison of the crisp and probabilistic approaches through experimentation on a benchmark dataset of human surveillance videos.
\end{abstract}


\section{Introduction}

Systems for event recognition --- `event pattern matching', in the terminology of \cite{luckhamBook} --- accept as input streams of sensor data in order to identify composite events of interest, that is, collections of events that satisfy some pattern. Consider, for example, the recognition of attacks on nodes of a computer network given the TCP/IP messages, the recognition of suspicious trader behaviour given the transactions in a financial market, the recognition of whale songs given a stream of whale sounds, and the recognition of human activities given multimedia content from surveillance cameras.

A common approach to event recognition separates low-level from high-level recognition. In the case of human activity recognition, the output of the former type of recognition is a set of activities taking place in a short period of time: `short-term activities' (STA). The output of the latter type of recognition is a set of  `long-term activities' (LTA), which are temporal combinations of STA. We focus on high-level recognition.

We define a set of LTA of interest, such as `fighting' and `meeting', as temporal combinations of STA such as `walking', `running', and `inactive' (standing still) using a logic programming (Prolog) implementation of the Event Calculus (EC) \cite{kowalski86}. We employ  EC to express
the temporal constraints on a set of STA that, if satisfied, lead to the recognition of a LTA.

In earlier work (Artikis, Sergot and Paliouras~\citeyear{artikis10EIMM}) we identified various types of uncertainty that exist in activity recognition, such as erroneous STA detection. To address this issue, we extend our work by presenting Prob-EC, an EC dialect suitable for probabilistic activity recognition. Prob-EC operates on the state-of-the-art probabilistic logic programming framework ProbLog \cite{kimmig11}. 
Prob-EC, therefore, may operate in settings where STA occurrences are assigned a confidence value by the underlying low-level tracking system (such as, for example, a probabilistic classifier).
We present extensive experimental evaluation of Prob-EC on a benchmark activity recognition dataset.
The evaluation demonstrates the conditions in which Prob-EC outperforms our previous EC dialect --- Crisp-EC. Prob-EC is the first EC dialect able to deal with uncertainty in the input STA. Moreover, this is the first approach that thoroughly evaluates EC in a probabilistic framework. 
The full code of Prob-EC, along with the dataset on which experimentation is performed, is available upon request.

The remainder of the paper is organised as follows. In the following section we set our work in context. In Section \ref{sec:ec} we present Crisp-EC while in Sections \ref{sec:stbr} and \ref{sec:ltbr} we describe, respectively, the dataset on which we perform activity recognition and the corresponding knowledge base of LTA definitions. Section \ref{sec:ProbLog} briefly introduces ProbLog while Section \ref{sec:EC_in_ProbLog} describes Prob-EC. Our experimental results, including the comparison between Prob-EC and Crisp-EC, are presented in Section \ref{sec:LTAR_EC_results}. Finally, in Section \ref{sec:summ} we summarise our observations and outline directions for further research.

\section{Related Work}\label{sec:relWork}

Numerous recognition systems have been proposed in the literature \cite{cugola11}. In this section we focus on long-term activity (LTA) recognition systems that, similar to our approach, exhibit a formal, declarative semantics. 

A fair amount of recognition systems is logic-based. Notable approaches include the Chronicle Recognition System \cite{dousson07} and the hierarchical event representation of \cite{shah07AIJ}. A recent review of logic-based recognition systems may be found in (Artikis, Skarlatidis et al.~\citeyear{artikisKER}). These systems are common in that they employ logic-based methods for representation and inference, but are unable to handle noise.

Shet et al.~\citeyear{davis05AVSS,davis07CVPR} have presented a logic programming approach to activity recognition which touches upon the issue of data coming from noisy sensors. In that work, LTA concerning theft, entry violation, unattended packages, and so on, have been defined. Within their activity recognition system, Shet and colleagues have incorporated mechanisms for reasoning over rules and facts that have an uncertainty value attached. Uncertainty in rules corresponds to a measure of rule reliability. On the other hand, uncertainty in facts represents the detection probabilities of the short-term activities (STA). In the VidMAP system \cite{davis05AVSS}, a mid-level module which generates Prolog facts automatically filters out data that a low-level image processing module has misclassified (such as a tree mistaken for a human). Shet and colleagues have noted of the filtering carried out by this module that `...it does so by observing whether or not the object has been persistently tracked' \cite[p.
~2]{davis05AVSS}. 
In \cite{davis07CVPR}, the authors  use an algebraic data structure known as a bilattice \cite{ginsberg90} to detect human entities based on uncertain output of part-based detectors, such as head or leg detectors. The bilattice structure associates every STA or LTA with two uncertainty values, one encoding available information and the other encoding confidence. The more confident information is provided, the more probable the respective LTA becomes.

A ProbLog-based method for robotic action recognition is proposed in \cite{Moldovan12a}. The method employs a relational extension of the \emph{affordance models} \cite{gibson1979ecological} in order to represent multi-object interactions in a scene. Affordances can model the relations between objects, actions (that is, pre-programmed robotic arm movements) and effects of actions. In contrast to a standard propositional Bayesian Network implementation of an affordance model, the method can scale to multiple object interactions in a scene without the need of retraining. However, the proposed method does not include temporal representation and reasoning.

 
%
%
Probabilistic graphical models have been applied on a variety of activity recognition applications where uncertainty exists. Activity recognition requires processing streams of timestamped STA and, therefore, numerous activity recognition methods are based on sequential variants of probabilistic graphical models, such as Hidden Markov Models \cite{rabiner1986HMM}, Dynamic Bayesian Networks \cite{murphy2002DBN} and Conditional Random Fields \cite{LaffertyMP01CRF}. Compared to logic-based methods, graphical models can naturally handle uncertainty but their propositional structure provides limited representation capabilities. To model LTA that involve a large number of relations among STA, such as interactions between multiple persons and/or objects, the structure of the model may become prohibitively large and complex. To overcome such limitations, these models have been extended in order to support more complex relations. Examples of such extensions include representing interactions involving multiple domain 
objects \cite{BrandOP97,gong2003,wu2007joint,vail2007conditional}, capturing long-term dependencies between states \cite{hongeng2003large}, as well as describing a hierarchical composition of activities \cite{natarajan2007hierarchical,liao2007hierarchical}. However, the lack of a formal representation language makes the definition of complex LTA complicated and the integration of domain background knowledge very hard. 

%
%
Markov Logic Networks (MLN) \cite{mln2006} have also been used for representing uncertainty in activity recognition. MLN employ first-order logic representation, where each formula may be associated with a weight, indicating the confidence we have on the formula. The knowledge base of weighted formulas is translated into a Markov network where probabilistic inference is performed. The approach of \cite{BiswasTF07}, for example, uses MLN to recognise LTA given the STA that have been observed by low-level classifiers. A more expressive approach that can represent persistent and concurrent LTA, as well as their starting and ending time-points, is proposed in \cite{HelaouiNS11}. The method in \cite{sadilek2012} employs hybrid-MLN  \cite{wang2008hybrid} in order to recognise successful and failed interactions between multiple humans using noisy location data. Similar to pure MLN-based methods, the knowledge base is composed by LTA definitions. Furthermore, hybrid formulas aiming to remove the noise from the 
location data are also included. Hybrid formulas are defined as normal formulas, but their weights are also associated with a real-valued function, such as the distance of two persons. As a result, the confidence of the formula is defined by both its weight and function. Although these methods incorporate first-order logic representation, the presented LTA definitions have a limited temporal representation. 

A method that uses interval-based temporal relations is proposed in \cite{morariu11}. The aim of the method is to determine the most consistent sequence of LTA based on the observations of low-level classifiers. Similar to \cite{davisMLN2008,davis2010}, the method uses MLN to express LTA. In contrast to \cite{davisMLN2008,davis2010}, it employs temporal relations based on Allen's Interval Algebra (IA) \cite{Allen83}. In order to avoid the combinatorial explosion of possible intervals that IA may produce, a bottom-up process eliminates the unlikely LTA hypotheses. In \cite{brendel2011PEL,selman2011PEL} a probabilistic extension of Event Logic \cite{siskind2001EL} is proposed in order to perform interval-based activity recognition. Similar to MLN, the method defines a probabilistic model from a set of domain-specific weighted LTA. However, the Event Logic representation avoids the enumeration of all possible interval relations.

%
%

The main difference of our approach with respect to the aforementioned lines of work concerns the fact that we use the Event Calculus (EC) for temporal representation. EC has built-in rules for complex temporal representation, including the formalisation of inertia, which help considerably the system designer in developing activity definitions. With the use of EC one may develop intuitive, succinct activity definitions, facilitating the interaction between activity definition developer and domain expert, and allowing for code maintenance. Furthermore, being logic programming-based, EC has direct routes to the ProbLog probabilistic logic programming framework. To the best of our knowledge, the probabilistic EC dialect presented in this paper, Prob-EC, is the first EC dialect able to deal with uncertainty in STA detection.

A MLN-based approach that is complementary to our work is that of \cite{anskarlRuleML}, which introduces a probabilistic EC dialect based on MLN. This dialect and Prob-EC tackle the problem of probabilistic inference from different viewpoints. Prob-EC handles noise in the input stream, represented as detection probabilities of the STA. The MLN-based EC dialect, on the other hand, emphasises uncertainty in activity definitions in the form of rule weights.

ProbLog and MLN are closely related. A notable difference between them is that MLN, as an extension of first-order logic, are not bound by the closed-world assumption. There exists a body of work that investigates the connection between the two frameworks. \cite{foproblog}, for example, developed an extension of ProbLog which is able to handle first-order formulas with weighted constraints. Fierens et. al \citeyear{weightedcnfs} converted probabilistic logic programs to ground MLN and then used state-of-the-art MLN inference algorithms to perform inference on the transformed programs. 

\begin{table}[t]
\begin{center}
\renewcommand{\arraystretch}{0.9}
\setlength\tabcolsep{3pt}
\begin{tabular}{ll}
\hline\noalign{\smallskip}
\multicolumn{1}{c}{\textbf{Predicate}} & \multicolumn{1}{c}{\textbf{Meaning}} 
\\
\noalign{\smallskip}
\hline
\noalign{\smallskip}
$\happens(E,\ T)$ & Event $E$ is occurring at time $T$  \\[2pt]

$\initially( F \val V )$ & The value of fluent $F$ is $V$ at time $0$  \\[2pt]

$\holdsAt(F \val V,\ T)$ & The value of fluent $F$ is $V$ at time $T$ \\[2pt]

$\initiatedAt( F \val V,\ T)$ & At time $T$ a period of time for which $F\val V$ is initiated  \\[2pt]

$\terminatedAt(F \val V,\ T)$ & At time $T$ a period of time for which $F\val V$ is terminated\\ \hline
\end{tabular}
\end{center}
\caption{Main Predicates of Crisp-EC.}\label{tbl:ec}
\end{table}

\section{The Event Calculus} \label{sec:ec}

Our LTA recognition system is based on a logic programming (Prolog) implementation of an EC dialect. EC, introduced by Kowalski and Sergot \citeyear{kowalski86}, is a many-sorted, first-order predicate calculus for representing and reasoning about events and their effects. For the dialect presented here --- Crisp-EC --- the time model is linear and includes integers. Where $F$ is a \emph{fluent} --- a property that is allowed to have different values at different points in time --- the term $F \val V$ denotes that fluent $F$ has value $V$. Boolean fluents are a special case in which the possible values are \true\ and \false. 
Informally, $F \val V$ holds at a particular time-point if $F \val V$ has been \emph{initiated} by an
event at some earlier time-point, and not \emph{terminated} by another event in the meantime --- law of inertia.

We represent STA as events and LTA as fluents. In this way, we can express the conditions in which the occurrence of a STA initiates or terminates a LTA.

An \emph{event description} in Crisp-EC includes rules that define, among other things, the event occurrences (with the use of the \happens\ predicate), the effects of events (with the use of the \initiatedAt\ and
\terminatedAt\ predicates), and the values of the fluents (with the use of the \initially\ and \holdsAt\ predicates). Table \ref{tbl:ec} summarises the main predicates of Crisp-EC. Variables, starting with an upper-case letter, are assumed to be universally quantified unless otherwise indicated. Predicates, function symbols and constants start with a lower-case letter.

The domain-independent rules for \holdsAt\ can be written in the following form:
\begin{align}
 &
 \label{eq:ec-holdsAt-init}
 \begin{mysplit}
 \holdsAt(F\val V,\ T) \leftarrow \\
 \quad   \initially(F\val V),\\
 \quad   \nbf\ \broken(F\val V,\ 0,\ T)
 \end{mysplit}
\end{align}
\begin{align}
&
\label{eq:ec-holdsAt}
\begin{mysplit}
\holdsAt(F\val V,\ T) \leftarrow \\
\quad   \initiatedAt(F\val V,\ T_s),\\
\quad T_s < T, \\ 
\quad \nbf\ \broken(F\val V,\ T_s,\ T)
\end{mysplit}\\
& \label{eq:ec-broken-terminated}
\begin{mysplit}
\broken(F\val V,\, T_s,\, T) \leftarrow \\
\quad   \terminatedAt(F\val V,\ T_f),\\ \quad T_s < T_f < T
\end{mysplit}\\
& \label{eq:ec-broken-initiated}
\begin{mysplit}
\broken(F\val V_1,\ T_s,\ T) \leftarrow \\
\quad  \initiatedAt(F\val V_2,\ T_f),\\ \quad V_1 \neq V_2, \\ \quad T_s < T_f < T
\end{mysplit}
\end{align}
\nbf\ represents `negation by failure', which provides a form of default persistence --- inertia --- of fluents. 
According to rule \eqref{eq:ec-holdsAt-init}, $\mathit{F\val V}$ holds at time-point $\mathit{T}$ if $\mathit{F\val V}$ held initially and has not been \broken\ since.
According to rule \eqref{eq:ec-holdsAt}, $\mathit{F\val V}$ holds at time-point $\mathit{T}$ if the fluent $\mathit{F}$ has been initiated to value $\mathit{V}$ at an earlier time $\mathit{T_s}$, and has not been \broken\ since.  According to rule \eqref{eq:ec-broken-terminated}, a period of time for which $F\val V$ holds is \broken\ at $T_f$ if $F\val V$ is terminated at $T_f$. Rule \eqref{eq:ec-broken-initiated} dictates that if $F \val V_2$ is initiated at $T_f$ then effectively $F \val V_1$ is terminated at $T_f$, for all other possible values $V_1$ of $F$. Rule \eqref{eq:ec-broken-initiated} therefore ensures that a fluent cannot have more than one value at any time. We do not insist that a fluent must have a value at every time-point. In Crisp-EC there is a difference between initiating a Boolean fluent $F \val \false$ and terminating $F \val \true$: the first implies, but is not implied by, the second.

According to rules \eqref{eq:ec-holdsAt}--\eqref{eq:ec-broken-initiated}, $F\val V$ does not hold at the time it was initiated, while it holds at the time it was terminated.

The definitions of \initiatedAt\ and \terminatedAt\ are domain-specific. One common form of rule for \initiatedAt\ is the following:
\begin{align}
& \label{eq:ec-initiatedAt-typical}
\begin{mysplit}
\initiatedAt(F\val V,\ T) \leftarrow \\
\quad  \happens(E,\ T), \\ \quad \mathit{Conditions}[T]
\end{mysplit}
\end{align}
where $\mathit{Conditions}[T]$ is some set of further conditions referring to time-point $T$. \linebreak\terminatedAt\ rules are handled similarly. 
Note that in this EC formulation, \linebreak$\initiatedAt(F\val V,\ T)$ does not necessarily imply that $F \notval V$ at $T$. Similarly, \linebreak$\terminatedAt(F\val V,\ T)$ does not necessarily imply that $F\val V$ at $T$. Suppose, for example, that $F\val V$ is initiated at time-point $20$ and terminated at time-point $30$ and that there are no other time-points at which it is initiated or terminated. Then $F\val V$ holds at all time-points $T$ such that $20 < T \leq 30$.  Suppose now that $F\val V$ is initiated at time-points $10$ and  $20$ and terminated at time-point $30$ (and at no other time-points). Then $F\val V$ holds at all $T$ such that $10 < T \leq 30$. And suppose finally that $F\val V$ is initiated at time-points $10$ and  $20$ and terminated at time-points $25$ and  $30$ (and at no other time-points). In that case $F\val V$ holds at all  $T$ such that $10 < T \leq 25$. 
In Section \ref{sec:ltbr} we illustrate the use of \initiatedAt\ and \terminatedAt\ rules for expressing LTA definitions. 


\section{Short-Term Activities}\label{sec:stbr}

We use the first dataset of the CAVIAR project\footnote{\urlBiBTeX{http://homepages.inf.ed.ac.uk/rbf/CAVIAR/}} to perform LTA recognition. This dataset includes 28 surveillance videos of a public space. The videos are staged --- actors walk around, sit down, meet one another, leave objects behind, fight, and so on. Each video has been manually annotated by the CAVIAR team in order to provide the ground truth for both STA and LTA. For this set of experiments, the input to our recognition system is:

\begin{enumerate}
 \item[(i)] The STA \walking, \running, \activeb\ (non-abrupt body movement in the same position) and \inactive\ (standing still), together with their time-stamps, that is, the video frame in which that STA took place. 
The original CAVIAR dictionary does not include a STA for `abrupt motion'. Our preliminary experiments with this dataset showed that the absence of such a STA compromises the recognition accuracy of some LTA. `Abrupt motion' is a form of STA that is detected by some state-of-the-art detection systems, for example \cite{kosmo08setn}, but not by the CAVIAR systems. Accordingly, we modified the CAVIAR dataset by introducing a STA for `abrupt motion': we manually edited the annotation of the CAVIAR videos by changing, when necessary, the label of a STA to \abrupt.
The STA \abrupt, \walking, \running, \activeb\ and \inactive\ are mutually exclusive and represented by means of the \linebreak\happens\ predicate. For example, $\mathit{\happens(active(id_6), 80)}$ expresses that $\mathit{id_6}$ displayed `active' bodily movement at time-point $\mathit{80}$. STA are represented as instantaneous events in EC in order to use the \initiatedAt\ and \terminatedAt\ predicates to express the conditions in which these activities initiate and terminate a LTA.  

\item[(ii)] The coordinates of the tracked people and objects as pixel positions at each time-point, as well as their orientation. The coordinates are represented with the use of the \holdsAt\ predicate. $\mathit{\holdsAt(coord(id_2)\val (14, 55),\ 10600)}$, for example, expresses that the coordinates of $\mathit{id_2}$ are $\mathit{(14, 55)}$ at time-point $\mathit{10600}$. Orientation is also encoded using the \holdsAt\ predicate. For instance, $\mathit{\holdsAt(orientation(id_2)\val 120,\ 10600)}$  expresses that, in the two-dimensional projection of the video, the same person was forming a 120$^{\circ}$ angle with the x-axis at the same time-point. This type of information is necessary for computing the distance between two entities as well as the direction to which a person might be headed.

\item[(iii)] The first and the last time a person or object is tracked (`appears' and `disappears'). This type of input is represented using the \happens\ predicate. For example,  $\mathit{\happens(appear(id_{10}),\  300 )}$ expresses that $\mathit{id_{10}}$ is first tracked at time-point $\mathit{300}$. 
\end{enumerate}

Given such input, Crisp-EC recognises the following LTA: a person leaving an object, people meeting, moving together, or fighting. Long-term activities are represented as EC fluents. For instance, $\mathit{\holdsAt( moving(id_1, id_3)\val\true, 140 )}$ states that $id_1$ was moving together with $id_3$ at time-point $\mathit{140}$.

LTA recognition is based on a manually developed knowledge base of LTA definitions expressed in terms of \initiatedAt\ and \terminatedAt. In the next section, we present example definition fragments of the LTA knowledge base.
The complete code is available upon request.

\section{Long-Term Activity Definitions}\label{sec:ltbr}

 
The `leaving an object' activity is defined as follows:
\begin{align}
& \label{eq:leaving-inactive-init}
\begin{mysplit}
\initiatedAt(\mathit{leaving\_object(P,\ Obj)\val\true,\ T} ) \leftarrow \\
\quad\happens( \mathit{appear(Obj) ,\ T} ), \\
\quad\happens( \mathit{inactive(Obj),\ T} ), \\
\quad\holdsAt( \mathit{close(P,\ Obj,\ 30)\val\true,\ T} ), \\
\quad\holdsAt( \mathit{person(P)\val\true,\ T} )
\end{mysplit}\\
& \label{eq:leaving-exit-term}
\begin{mysplit}
\terminatedAt( \mathit{leaving\_object(P,\ Obj)\val\true,\ T} ) \leftarrow \\
\quad\happens( \mathit{disappear(Obj),\ T} )
\end{mysplit}
\end{align}
In the CAVIAR videos an object carried by a person is not tracked --- only the person that carries it is tracked. The object will be tracked (`appear') if and only if the person leaves it somewhere. Moreover, objects (as opposed to persons) can only exhibit the \inactive\ STA. Accordingly, rule
\eqref{eq:leaving-inactive-init} expresses the conditions in which `leaving an object' is recognised. The fluent recording this activity, \leave$\mathit{(P, Obj)}$, becomes \true\ at time $T$ if $\mathit{Obj}$ `appears' at $T$, its STA at $T$ is `inactive', and there is a person $P$ `close' to
$\mathit{Obj}$ at $T$. 
The $\mathit{close(ID1, ID2, Threshold)}$ fluent expresses that the distance between $\mathit{ID1}$ and $\mathit{ID2}$ is at most $\mathit{Threshold}$
pixels. This fluent is defined as follows:
\begin{align}
& \label{eq:close}
\begin{mysplit}
\holdsAt(\mathit{close(ID1, ID2, Threshold) \val \true,\ T} ) \leftarrow \\ 
\quad \holdsAt(\mathit{distance(ID1,ID2) \val Dist, \ T} ), \\
\quad \mathit{Dist < Threshold}
\end{mysplit}
\end{align}
The distance between two tracked objects/people is computed as the Euclidean distance between their
coordinates in the two-dimensional projection of the video --- recall that the coordinates of each tracked entity are given as input to our system.

The 30 pixel distance threshold in rule \eqref{eq:leaving-inactive-init} was determined from an empirical analysis of the CAVIAR dataset.

An object that is picked up by someone is no longer tracked (it `disappears') which in turn terminates \leave\ --- see rule \eqref{eq:leaving-exit-term}.

In CAVIAR there is no explicit information that a tracked entity is a person or an inanimate object. Therefore, in our activity definitions we try to deduce whether a tracked entity is a person or an object given the detected STA. We defined the fluent $\mathit{person(P)}$ to have value \true\ if $P$ was active, walking, running or moved abruptly at some time-point since $P$ `appeared'.
\begin{align}
& \label{eq:person} 
\begin{mysplit}
\mathit {\initiatedAt(person(P)\val\true,\ T)} \leftarrow \\ 
\quad\happens( \mathit{active(P),\ T} )  \\ 
\mathit {\initiatedAt(person(P)\val\true,\ T)} \leftarrow \\ 
\quad\happens( \mathit{walking(P),\ T}\ )  \\ 
\mathit {\initiatedAt(person(P)\val\true,\ T)} \leftarrow \\ 
\quad\happens( \mathit{running(P),\ T} )  \\
\mathit {\initiatedAt(person(P)\val\true,\ T)} \leftarrow \\ 
\quad\happens( \mathit{abrupt(P),\ T} )  \\
\mathit{\terminatedAt(person(P)\val\true,\ T)} \leftarrow \\ 
\quad \mathit{\happens(disappear(P),\ T)} 
\end{mysplit}
\end{align}
The value of $\mathit{person(P)}$ is time-dependent because in CAVIAR the identifier $P$ of a tracked entity that `disappears'  (is no longer tracked) at some point  may be used later to refer to another entity that `appears' (becomes tracked), and that other entity may not necessarily be a person.
Note, finally, that rule \eqref{eq:leaving-inactive-init} incorporates a (reasonable) simplifying assumption,  that a person entity will never exhibit `inactive' activity at the moment it first `appears' (is tracked). If an entity is `inactive' at the moment it  `appears' it can be assumed to be an object, as in the first two conditions of rule \eqref{eq:leaving-inactive-init}.

In a similar way, we may express the definitions of other LTA. The use of EC, in combination with the full power of logic programming, allows us to express LTA definitions including complex temporal, spatial or other constraints. Below we present fragments of the remaining LTA definitions.
%


\meet\ (of two persons $\mathit{P_1}$ and $\mathit{P_2}$) is recognised when two people `interact': at least one of them is active or inactive, the other is not running or moving abruptly, and the distance between them is at most 25 pixels (all numeric constraints were determined from an empirical analysis of the dataset). In CAVIAR, this interaction phase can be seen as some form of greeting, such as a handshake. Rules \eqref{eq:meet-inactive-init} and \eqref{eq:meet-active-init} show the conditions in which \meet\ is initiated:
\begin{align}
& \label{eq:meet-inactive-init}
\begin{mysplit}
\initiatedAt( \mathit{meeting(P_1,\ P_2)\val\true,\ T} )\leftarrow \\
\quad \happens( \mathit{inactive(P_1),\ T} ),\\
\quad\holdsAt( \mathit{close(P_1,\ P_2,\ 25)\val\true,\ T} ),\\
\quad \holdsAt( \mathit{person(P_1)\val\true,\ T} ),\\
\quad \holdsAt( \mathit{person(P_2)\val\true,\ T} ),\\
\quad\nbf\ \happens( \mathit{running(P_2),\ T} ),\\
\quad\nbf\ \happens( \mathit{abrupt(P_2),\ T} ),\\
\quad\nbf\ \happens( \mathit{active(P_2),\ T} )
\end{mysplit}\\
& \label{eq:meet-active-init}
\begin{mysplit}
\initiatedAt( \mathit{meeting(P_1,\ P_2)\val\true,\ T} )\leftarrow \\
\quad \happens( \mathit{active(P_1),\ T} ),\\
\quad\holdsAt( \mathit{close(P_1,\ P_2,\ 25)\val\true,\ T} ),\\
\quad \holdsAt( \mathit{person(P_2)\val\true,\ T} ),\\
\quad\nbf\ \happens( \mathit{running(P_2),\ T} ),\\
\quad\nbf\ \happens( \mathit{abrupt(P_2),\ T} )
\end{mysplit}
\end{align}

\meet\ is terminated by a plethora of conditions, such as when one of the two people involved in the LTA starts running or `disappears'. 

In CAVIAR \meet\ may have multiple initiations: two people may be interacting for several video frames. Similarly, \meet\ may have multiple terminations. This is in contrast to \leave\ where there is a single initiation and a single termination --- for example, an object `appears' once before it `disappears'. In general, there is no fixed relation between the number of initiations and terminations of a fluent --- for example, a LTA may have multiple initiations and a single termination. In Section \ref{sec:ec} we described how Crisp-EC computes the time-points in which a fluent with one or more initiations and terminations holds.

The activity \move\ was defined in order to recognise whether two people are walking along together:
\begin{align}
& \label{eq:moving-init}
 \begin{mysplit}
 \initiatedAt \mathit{( moving(P_1,\ P_2)\val\true,\ T} )\leftarrow \\
\quad \mathit{\happens(walking(P_1), T),} \\
\quad \mathit{\happens(walking(P_2), T),} \\
\quad \mathit{\holdsAt( close(P_1,\ P_2,\ 34)\val\true,\ T )}, \\
\quad \mathit{\holdsAt( orientation(P_1)\val Or_1,\ T )}, \\
\quad \mathit{\holdsAt( orientation(P_1)\val Or_2,\ T )}, \\
\quad \mathit{|Or_1 - Or_2| < 45}
 \end{mysplit}
\end{align}
In order to recognise \move, both people involved have to be \walking\ while being close to each other. In addition, they have to be facing towards, more or less, the same direction (people walking in opposite directions are not assumed to be walking along together). This is accomplished by constraining their orientations so that they are, roughly, headed towards the same area while they are walking. 

LTA --- in contrast to STA --- are not mutually exclusive. For example, \meet\ may overlap with \move: two people interact and then start \move, that is, they walk while being \close\ to each other. In general, however, there is no fixed relationship between LTA.

\move\ is terminated when either person walks away from the other with respect to the predefined threshold of 34 pixels:
\begin{align}
& \label{eq:moving-term}
 \begin{mysplit}
 \terminatedAt \mathit{( moving(P_1,\ P_2)\val\true,\ T} )\leftarrow \\
\quad \mathit{\happens(walking(P_1), T),} \\
\quad \mathit{\holdsAt(close(P_1,\ P_2,\ 34)\val\false,\ T )} \\
 \end{mysplit}
\end{align}

Other termination conditions for \move\ include either person running away from the other, as well as either person `disappearing' from the scene.

%
The last definition concerns the \fight\ activity:
\begin{align}
& \label{eq:fighting-init-abrupt}
\begin{mysplit}
\initiatedAt \mathit{(fighting(P_1,\ P_2)\val\true,\ T})\leftarrow \\
\quad \mathit{\happens(abrupt(P_1),\ T),}\\
\quad \mathit{\holdsAt(close(P_1,\ P_2,\ 44)\val\true,\ T)}, \\
\quad \mathit{\nbf\ \happens(inactive(P_2),\ T)} 
\end{mysplit}
\end{align}
To recognise \fight, we require that both people are sufficiently close and at least one of them moves abruptly, while the other one is not \inactive, indicating that he ought to be participating in the fight somehow. \fight\ ceases to be recognised when either person involved in the activity starts walking away, running away, or exits the scene. 

\section{A Probabilistic Logic Programming Framework}\label{sec:ProbLog}

In earlier work (Artikis, Sergot and Paliouras~\citeyear{artikis10EIMM}) we identified various types of uncertainty that exist in activity recognition, such as erroneous STA detection. To address this issue, we ported our EC dialect into ProbLog \cite{kimmig11}, a probabilistic extension of the logic programming language Prolog. ProbLog differs from Prolog in that it allows for probabilistic facts, which are facts of the form $p_i::f_i$ where $p_{i}$ is a real number in the range [0, 1] and $f_{i}$ is a Prolog fact. If $f_{i}$ is not ground, then the probability $p_{i}$ is applied to all possible groundings of
$f_{i}$. Classic Prolog facts are silently given probability 1.

Probabilistic facts in a ProbLog program represent random variables. Furthermore, ProbLog makes an independence assumption on these variables. This means that a rule which is defined as a conjunction of $n$ of these probabilistic facts has a probability equal to the product of the probabilities of these facts. When a predicate appears in the head of more than one rule then its probability is computed by calculating the probability of the implicit disjunction created by the multiple rules. 
For example, for a predicate $\mathit{p}$ with two rules $\mathit{p \leftarrow l_1 }$ and $\mathit{p \leftarrow l_2, l_3}$,
the probability $\mathit{P(p)}$ is computed as follows:
\begin{align}
& \nonumber
\begin{mysplit}
P(p) \val P((p \leftarrow l_1) \vee (p \leftarrow l_2, l_3)) \val \\
\qquad \ \val  P(p \leftarrow l_1) + P(p \leftarrow l_2, l_3) - P((p \leftarrow l_1) \wedge (p \leftarrow l_2, l_3)) \val \\
\qquad \ \val P(l_1) + P(l_2) \times P(l_3) - P(l_1) \times P(l_2) \times P(l_3)
\end{mysplit}
\end{align}

Given the independence assumption, any subprogram $\mathit{L}$ has a probability equal to:
\begin{align}
& \label{eq:subprogram_prob}
\begin{mysplit}
\mathit{P(L) = \prod \limits_{f_i \in L} p_i \cdot \prod \limits_{f_i \notin L} (1 - p_i)} \\
\end{mysplit}
\end{align}
With the help of equation \eqref{eq:subprogram_prob}, one could compute the probability that a query $\mathit{q}$ holds in a ProbLog program --- \emph{success probability} --- by summing the probabilities of all subprograms that entail it:
\begin{align}
& \label{eq:successprob_def}
\begin{mysplit}
\mathit{P_s(q) = \sum \limits_{L \models q} P(L) }\\
\end{mysplit}
\end{align}
Computing the success probability through equation \eqref{eq:successprob_def}, however, is computationally infeasible for large programs, since it involves
summing through an exponential number of summands ($ 2^{|B_L|}$ different subprograms, where $\mathit{B_L}$ is the Herbrand Base). By combining equations \eqref{eq:subprogram_prob} and \eqref{eq:successprob_def} and eliminating redundant
terms, we end up with the following characterisation:
\begin{align}
& \label{eq:successprob_comp}
\begin{mysplit}
\mathit{P_s(q) = P(\bigvee \limits_{e \in Proofs(q)}\ \bigwedge \limits_{f_i \in e} f_i\ )}\\
\end{mysplit}
\end{align}
That is, the task of computing the success probability of a query $\mathit{q}$ is transformed into the task of computing the probability of the Disjunctive Normal Form (DNF) formula of equation \eqref{eq:successprob_comp}. Practically, equation \eqref{eq:successprob_comp} expresses that the success probability of query $q$ is equal to the probability that at least one of its proofs is sampled. This, unfortunately, is not a question of straightforwardly transforming the probability of the DNF to a sum of products. 
If we were to translate equation \eqref{eq:successprob_comp} to a sum of products, we would assume that all different proofs (conjunctions) are disjoint, meaning that they represent mutually exclusive possible worlds, which does not hold in the general case. In order to make the proofs disjoint, one would have to enhance every conjunction with negative literals, in order to exclude worlds whose probability has already been computed in previous conjunctions of the DNF. This problem is known as the \emph{disjoint-sum problem} and is known to be \#P-hard \cite{disjointsum}.

ProbLog's approach consists of using Binary Decision Diagrams (BDDs) \cite{bdd} to compactly represent the DNF of equation  \eqref{eq:successprob_comp}. A BDD is a binary decision tree with redundant nodes removed and isomorphic subtrees merged. The BDD nodes represent the probabilistic facts of the ProbLog program. Every node has a `positive' and `negative' outward edge, leading to either a child node or the special `true' or `false' terminal nodes. The positive outward edge of the BDD node is labelled with the probability of the respective probabilistic fact and the negative edge is labelled with the complement of that probability. Positive and negative edges represent distinct decisions on inclusion of the relevant fact in the currently sampled possible world; a positive edge signifies that the fact represented by its parent node is included in the sample with the labelled probability, whereas a negative edge signifies that the fact is not included in the sample with the complement of the same probability.
 Therefore, by following a path from the root node to the `true' terminal node, one could sample a conjunction of the DNF formula of equation \eqref{eq:successprob_comp}. The `negative' outward edges offer a compact representation of the negated literals required to enhance the DNF formula in order to make it represent a disjunction over disjoint conjunctions.
 
To summarise, ProbLog's inference follows three general steps. The first step is to gather all proofs of the query $\mathit{q}$ by scanning the Selective Linear Definite (SLD) tree of proofs and represent
them as the DNF formula of equation \eqref{eq:successprob_comp}. Afterwards, with the help of a built-in translation script, the DNF is translated to
a BDD. Finally, the probability of this BDD is computed recursively, starting from the root node and assuming a probability of 1 for the `true' terminal and 0 for the `false' terminal.

With the help of BDDs, ProbLog inference is able to scale to queries containing thousands of different proofs \cite{kimmig11}. ProbLog's efficiency was the driving force behind our decision to use this framework for activity recognition under uncertainty.


ProbLog has been fully integrated in the YAP Prolog system.\footnote{\url{http://www.dcc.fc.up.pt/~vsc/Yap/}} 
Further details, examples and code samples are available in \cite{kimmig11} and on the ProbLog website.\footnote{\url{http://dtai.cs.kuleuven.be/problog/}}

\section{The Event Calculus in ProbLog}\label{sec:EC_in_ProbLog}

In this section we present the necessary transformations in order to make Crisp-EC ProbLog-compatible --- the result is Prob-EC. We also explain the inference procedure of Prob-EC through two examples.

\subsection{Transformation}\label{subsec:transf}

To express our EC dialect in ProbLog we had to update our treatment of negation in order to allow for probabilistic atoms. 
To compute the complement of the probability of a probabilistic fact, ProbLog provides the built-in predicate \pnot. For any probabilistic fact $\mathit{p_i::f_i}$, we have that:
\begin{align}
& \label{eq:problog_not_expl}
\begin{mysplit}
\mathit{P_s(\pnot(f_i)) \val 1 - P_s(f_i) \val 1 - p_i}
\end{mysplit}
\end{align}
\pnot\ can only be used on facts that are part of the knowledge base. For facts that do not belong to the knowledge base, such as a STA that is not part of an input stream, \pnot\ fails silently and reports a probability of 0. Consequently, translating \nbf\ to \pnot, where  facts are used in our EC axioms,
would lead to undesirable behaviour. Consider, for example, rule \eqref{eq:fighting-init-abrupt} re-written by translating \nbf\ to \pnot:
\begin{align}
& \label{eq:fighting-init-abrupt-pnot}
\begin{mysplit}
\initiatedAt \mathit{( fighting(P_1,\ P_2)\val\true,\ T} )\leftarrow \\
\quad \mathit{\happens(abrupt(P_1),\ T),}\\
\quad \mathit{\holdsAt(close(P_1,\ P_2,\ 44)\val\true,\ T )}, \\
\quad \mathit{\pnot}\mathit{(\happens(inactive(P_2),\ T))} \\
\end{mysplit}\tag{\ref{eq:fighting-init-abrupt}$'$}
\end{align}
Furthermore, assume that $\mathit{id_1}$ and $\mathit{id_2}$ start \fight\ at some time-point $t$: they are both moving abruptly and their distance is less than 44 pixel positions. In this case, however, \fight\ will not be initiated at $t$ as the final condition of rule \eqref{eq:fighting-init-abrupt-pnot} will not be satisfied: there are no \inactive\ facts at $t$ and therefore \pnot\ will fail.
To overcome this issue, we defined the following predicate:
\begin{align}
& \label{eq:negate1-1}
\begin{mysplit}
\negate_1\mathit{(Fact) \leftarrow \nbf\ Fact}
\end{mysplit}\\
& \label{eq:negate1-2}
\begin{mysplit}
\negate_1\mathit{(Fact) \leftarrow \pnot(Fact)} 
\end{mysplit}
\end{align}
With the use of \negate$_1$ we produce a probability of 1 whenever the negated STA is not detected (see rule \eqref{eq:negate1-1}), but also produce the complement of the probability of the STA whenever it is detected (see rule \eqref{eq:negate1-2}). Note that in the latter case, that is, when a STA is detected with some probability, then `\nbf\ STA' fails.

Rule \eqref{eq:fighting-init-abrupt}, for instance, can now be written as follows:
\begin{align}
& \label{eq:fighting-init-abrupt-prob}
\begin{mysplit}
\initiatedAt \mathit{( fighting(P_1,\ P_2)\val\true,\ T} )\leftarrow \\
\quad \mathit{\happens(abrupt(P_1),\ T),}\\
\quad \mathit{\holdsAt(close(P_1,\ P_2,\ 44)\val\true,\ T )}, \\
\quad \mathit{\negate}_1\mathit{(\happens(inactive(P_2),\ T))} \\
\end{mysplit}\tag{\ref{eq:fighting-init-abrupt}$''$}
\end{align}
With rule \eqref{eq:fighting-init-abrupt-prob} we are able to produce a probability of 1 whenever \linebreak$\mathit{\happens(inactive(P_2),\ T)}$ is not part of our input, as well as produce the complement of its probability whenever it is part of our input with a probability value attached. The only case when \negate$_1$ will produce a probability of 0 is when the probability value of its argument is 1. This is the desired behaviour.

The built-in predicate \pneg\ is an extension of \pnot\ applicable to both probabilistic facts and derived atoms. For a derived atom \textit{r}, we have:
\begin{align}
& \label{eq:problog_neg_expl}
\begin{mysplit}
\mathit{P_s(\pneg(r)) \val 1 - P_s(r)}
\end{mysplit}
\end{align}
Similarly to \pnot, \pneg\ cannot be used on goals that are not inferable from the knowledge base. Consider the following example:
\begin{align}
& \label{eq:ec-holdsAt-pneg}
\begin{mysplit}
\holdsAt(F\val V,\ T) \leftarrow \\
\quad   \initiatedAt(F\val V,\ T_s),\\ 
\quad T_s < T, \\ 
\quad  \pneg(\broken(F\val V,\ T_s,\ T))
\end{mysplit}\tag{\ref{eq:ec-holdsAt}$'$}
\end{align}
This rule is produced by replacing \nbf\ by \pneg\ in rule \eqref{eq:ec-holdsAt}. Assume that $F\val V$ has been initiated with some probability $p$ at $t{-}1$ and not \broken\ in the meantime. In this case, \holdsAt$\mathit{(F\val V,\ t)}$ cannot be proved because the last condition of rule \eqref{eq:ec-holdsAt-pneg} fails.
To overcome this issue, we defined the \negate$_2$ predicate:
\begin{align}
& \label{eq:negate2-1}
\begin{mysplit}
\negate_2\mathit{(Goal) \leftarrow \nbf\ Goal}
\end{mysplit}\\
& \label{eq:negate2-2}
\begin{mysplit}
\negate_2\mathit{(Goal) \leftarrow \pneg(Goal)} 
\end{mysplit}
\end{align}
With the use of \negate$_2$ we produce a probability of 1 whenever the negated goal is not inferable (see rule \eqref{eq:negate2-1}), but also produce the complement of the probability of the goal whenever it is inferable (see rule \eqref{eq:negate2-2}). 
Rule \eqref{eq:ec-holdsAt}, for instance, can now be re-written as follows:
\begin{align}
& \label{eq:ec-holdsAt-negate2}
\begin{mysplit}
\holdsAt(F\val V,\ T) \leftarrow \\
\quad   \initiatedAt(F\val V,\ T_s),\\ 
\quad T_s < T, \\ 
\quad  \negate_2(\broken(F\val V,\ T_s,\ T))
\end{mysplit}\tag{\ref{eq:ec-holdsAt}$''$}
\end{align}
If $F\val V$ is not \broken\ then the last condition of rule \eqref{eq:ec-holdsAt-negate2} will have probability 1 and, therefore, the probability that $F\val V$ will depend entirely on the probability of its initiation conditions. 
If, however, $F\val V$ is \broken\ with some probability $p$, then the probability that  $F\val V$ will be equal to the product of the probability of its initiation conditions and $1{-}p$. This is the desired behaviour.

\subsection{Inference}\label{subsec:inference}

To illustrate the inference procedure of Prob-EC we use two LTA from CAVIAR, one with multiple initiations and terminations, and one with a single initiation and a single termination.

\begin{figure}[t]
\centerline{
  \mbox{\includegraphics[width=\textwidth]{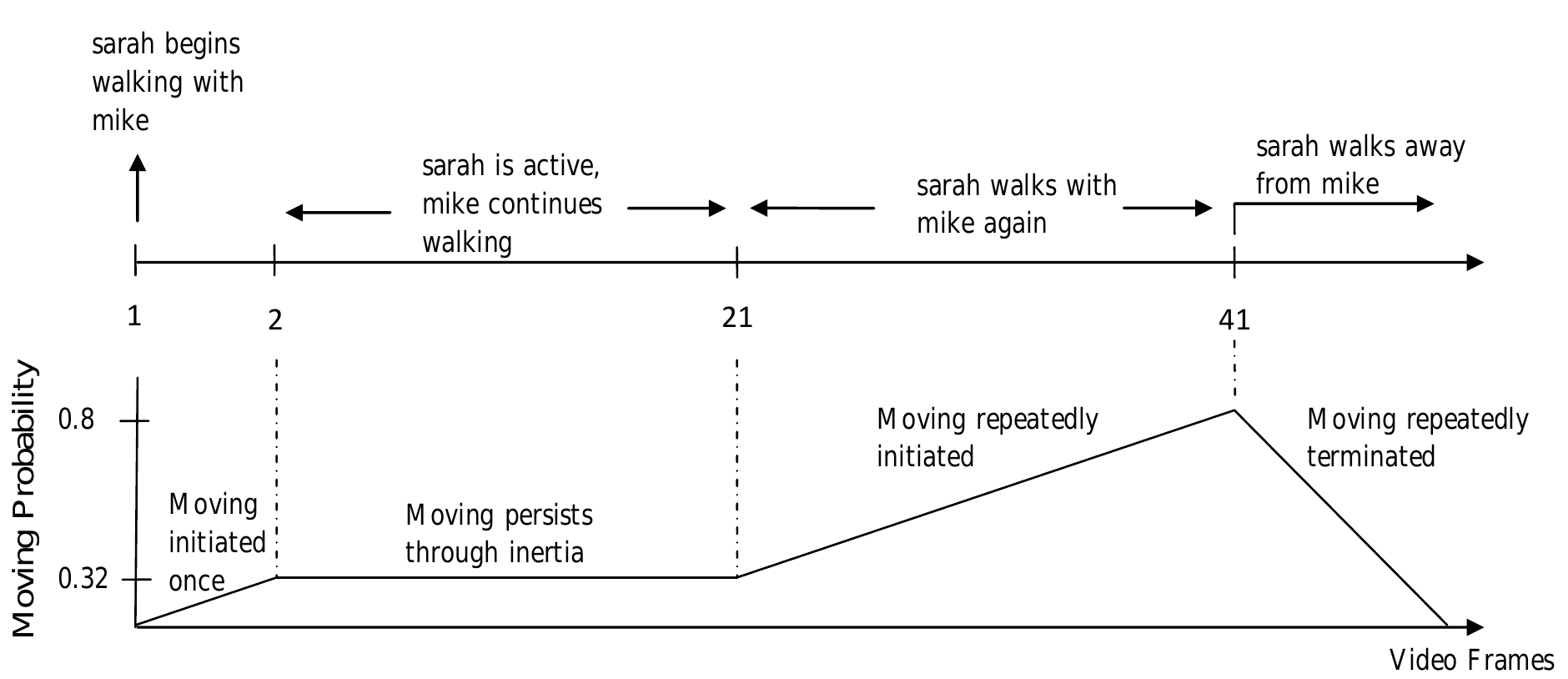}}}
\caption{Example of LTA probability fluctuation in the presence of multiple initiations and terminations.}
\label{fig:mikeandsarah}
\end{figure}

\subsubsection{Multiple Initiations and Terminations}\label{sec:multiple-initiations}

Suppose that \mike\ and \sarah\ are engaging in a `moving' activity for a number of video frames --- see Figure \ref{fig:mikeandsarah}. The activity is first initiated at frame number 1, when both \mike\ and \sarah\ start \walking. At frame 2 \sarah\ stops \walking\ (\walking\ is required by rule \eqref{eq:moving-init} to initiate \move). She instead displays \activeb\ body movement. Furthermore, \mike\ continues \walking\ but does not move far enough from her to terminate \move. At frame 21 \sarah\ resumes \walking, once again initiating \move. At frame 41 \sarah\ continues \walking, but \mike\ is \inactive\ and is left behind --- \sarah\ and \mike\ are no longer close enough to each other, which triggers the termination condition \eqref{eq:moving-term} of \move\ (`walk away'). For simplicity, let all information pertaining to orientation and coordinates be crisp (probability of 1). Moreover, assume that the STA \walking, \activeb\ and \inactive\ have probabilities attached, as follows: 

\begin{flushleft}
\begin{math}
\mathit{0.70::\happens(walking(\mike),\ 1)} \linebreak
\mathit{0.46::\happens(walking(\sarah),\ 1)} \linebreak
\mathit{0.73::\happens(walking(\mike),\ 2)} \linebreak
\mathit{0.55::\happens(active(\sarah),\ 2)} \linebreak
\ldots \linebreak
\mathit{0.69::\happens(walking(\mike),\ 21)} \linebreak
\mathit{0.58::\happens(walking(\sarah),\ 21)} \linebreak
\ldots \linebreak
\mathit{0.18::\happens(inactive(\mike),\ 41)} \linebreak
\mathit{0.32::\happens(walking(\sarah),\ 41)} \linebreak
\ldots 
\end{math}
\end{flushleft}
At frame 2, the query $\mathit{\holdsAt(moving(\mike,\ \sarah) \val \true,\ 2)}$ has a probability equal to the probability of the initiation condition of frame 1, which, according to rule \eqref{eq:moving-init}, and given that all coordinate and orientation-related information are crisply recognised, is the product of the  probabilities that both \mike\ and \sarah\ are \walking, that is, 0.70${\times}$0.46$\val$0.322. This is visualised at the far left of Figure \ref{fig:mikeandsarah}, where the LTA's probability jumps from 0 to 0.322. From frame 2 to frame 20 no initiation or termination conditions  are fired and the probability of \move\ remains unchanged. This occurs due to the law of inertia and is depicted graphically by the horizontal line between frames 2 and 20. At frame 21, \sarah\ starts walking alongside \mike\ again. Consequently, at frame 22, the query $\mathit{\holdsAt(moving(\mike,\ \sarah) \val \true,\ 22)}$ has two initiation conditions to consider, one fired at frame 1 and one at frame 21. 
This occurs because rule \eqref{eq:ec-holdsAt-negate2} searches over all time-points between time-point 0 and the current time-point for initiation conditions, finding both the condition fired at frame 1 and the one fired at frame 21. 

\begin{figure}[t]
\centering
\includegraphics[width=0.75\textwidth]{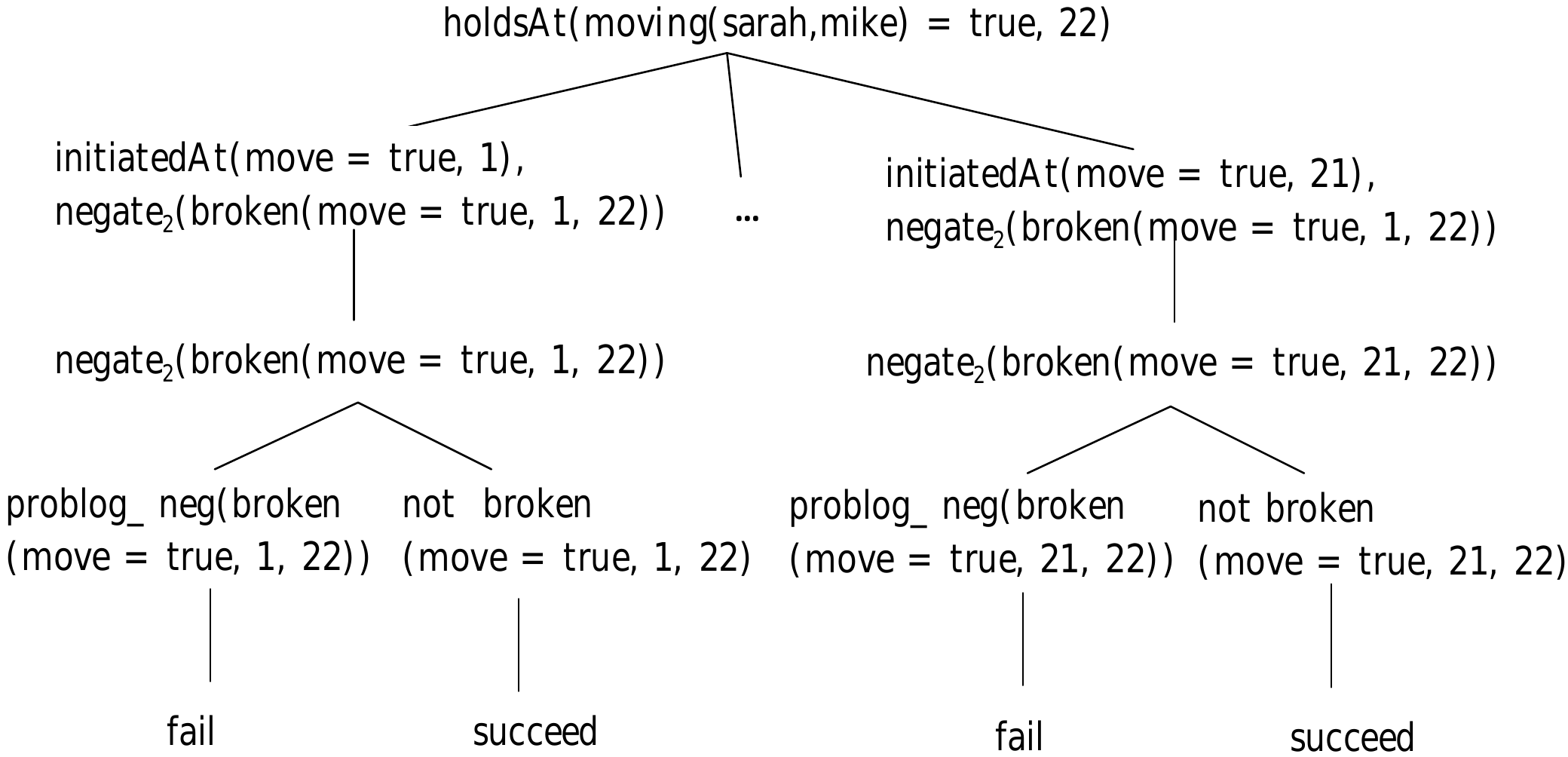}
\caption{SLD tree for query $\mathit{\holdsAt(moving(\mike,\ \sarah) \val \true,\ 22)}$. }
\label{fig:SLDTree}
\end{figure}

As mentioned in the previous section, ProbLog computes the probability of a query by first scanning the entire SLD tree of the query. Figure \ref{fig:SLDTree} depicts a fragment of the SLD tree for the query $\mathit{\holdsAt(moving(\mike,\ \sarah) \val \true,\ 22)}$. Then ProbLog represents these proofs as a DNF formula. In our case, the DNF is the following:
\begin{align}
& \label{eq:sarahmikednf}
\begin{mysplit}
\underbrace{\initiatedAt(\mathit{moving(\sarah, \mike) \val \true, 1)}}_\text{$\mathit{init_1}$} \bigvee \underbrace{\initiatedAt(\mathit{moving(\sarah, \mike) \val \true, 21)}}_\text{$\mathit{init_{21}}$}
\end{mysplit}
\end{align}
We have simplified the representation by omitting the two relevant `negate$_2$' clauses since, as can be seen in Figure \ref{fig:SLDTree}, they are both provable through negation as failure and therefore have a probability of 1. This occurs because no termination conditions for \move\ have been fired between frames 1 and 22.

Up to frame 22, there exist two initiation conditions for the \move\ LTA, $\mathit{init_1}$ and $\mathit{init_{21}}$ (see formula \eqref{eq:sarahmikednf}). In the general case, there may exist many more initiation conditions in the interval between the start of the video and the examined video frame. In addition, for every initiation condition, rule \eqref{eq:ec-holdsAt-negate2} will check whether the LTA has been terminated by examining the interval between the initiation and the current video frame, repeating the process at the next video frame. This leads to numerous redundant computations. We overcame this problem by implementing an elementary caching technique according to which the probability of $\mathit{\holdsAt(F \val V, T{-}1)}$ is stored in memory and, therefore, $\mathit{\holdsAt(F \val V, T)}$ simply checks to see how the initiation or termination conditions (if any) fired at time-point $\mathit{T{-}1}$ affect this probability. This technique operates under the assumption that the activity 
recognition system receives the video frames in a temporally sorted manner --- this assumption holds in CAVIAR. 

The next step of ProbLog inference involves translating the DNF into a BDD. However, our example is simple enough to allow us to perform manual calculations, as there exist only two proofs for \holdsAt, which are easy to disjoin. The probability of DNF formula \eqref{eq:sarahmikednf} can be computed as the probability of a disjunction of two elements, as explained in Section \ref{sec:ProbLog}:
\begin{multline*}
P(init_1 \vee init_{21}) \val P(init_1) + P(init_{21}) - P(init_1 \wedge \ init_{21}) \val \\
\val 0.70 \times 0.46 + 0.69 \times 0.58 -  0.7 \times 0.46 \times 0.69 \times 0.58 \val 0.593
\end{multline*} 
The probability that \mike\ and \sarah\ are \move\ at frame 22 has increased, owing to the presence of the additional initiation condition of frame 21. This is one of the characteristics of Prob-EC: the continuous presence of initiation conditions of a particular LTA causes an increase of the probability of the LTA. This behaviour is consistent with our intuition: given continuous indication that an activity has (possibly) occurred, we are more inclined to agree that it has indeed taken place, even if the confidence of every individual indication is low. For this reason, from frame 22 up to and including frame 41, the probability of \move\ increases, as is visible in Figure \ref{fig:mikeandsarah}. In this example, at frame 41 the activity's probability has escalated to around 0.8.

At frame 42, Prob-EC has to take into consideration the termination condition that was fired at frame 41. This termination condition, corresponding to rule \eqref{eq:moving-term}, is also probabilistic: it bears the probability that \sarah\ walked away from \mike, which, according to rule \eqref{eq:moving-term} and the fact that \close\ is crisply detected, is equal to the probability of the \walking\ STA itself, which is 0.32. Therefore, when estimating the probability that, at frame 42, \mike\ and \sarah\ are still \move\ together, we have to incorporate the probability of all possible worlds in which \sarah\ did not, in fact, walk away from \mike. The probability of these worlds is computed by the use of \negate$_2$ in rule \eqref{eq:ec-holdsAt-negate2} and is equal to 1${-}$0.32$\val$0.68. Consequently, the probability that \mike\ and \sarah\ are still \move\ together at frame 42 is 0.8${\times}$0.68$\val$0.544. Similarly to the steady probability increase given continuous initiation conditions, when 
faced with subsequent termination conditions, the probability of the LTA will steadily decrease. The slope of the descent (ascent)  is defined by the probability of the termination (initiation)  conditions involved. For this example, we assume that \sarah\ keeps walking away from \mike\ until the end of the video, causing the LTA's probability to approximate 0, as shown at the far right of Figure \ref{fig:mikeandsarah}. 

\begin{figure}[t]
\centering
\includegraphics[width=.7\textwidth]{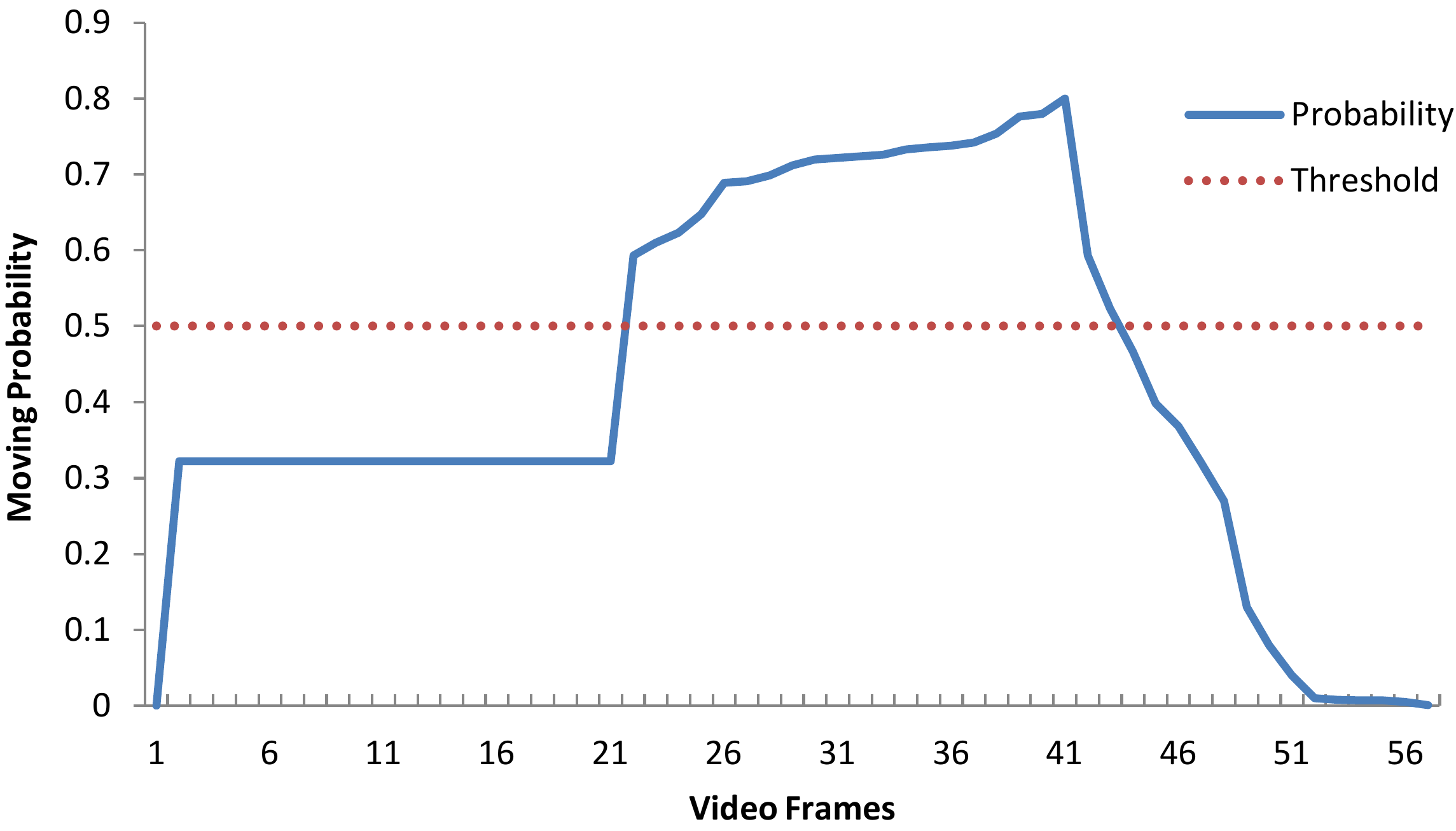}
\caption{ProbLog output concerning \holdsAt(\move(\mike, \sarah) $\val$ \true, T) for various video frames T. }
\label{fig:sarah_mike_plot}
\end{figure}

Figure \ref{fig:sarah_mike_plot} shows the precise ProbLog output for our example. After an abrupt jump from 0 to 0.322, the probability remains stable between frames 2 and 21, indicating that for this period of time the LTA persists through the law of inertia. Between frames 22 and 41, the LTA's probability monotonically increases, reflecting the repeated initiations of \move\ that occur during that time. After frame 41, it decreases, reflecting the repeated termination conditions occurring at the same time period. The dashed horizontal line at probability 0.5 represents the recognition threshold that we use to discern between LTA positives that we consider to be trustworthy enough --- these are the LTA recognitions --- and those that we do not. The choice of a 0.5 threshold was made simply to provide a concrete illustration --- other thresholds could have alternatively been used in this example.

\subsubsection{Single Initiation and Termination}\label{sec:single-initiation}



\begin{figure}[t]
\centering
\includegraphics[width=\textwidth]{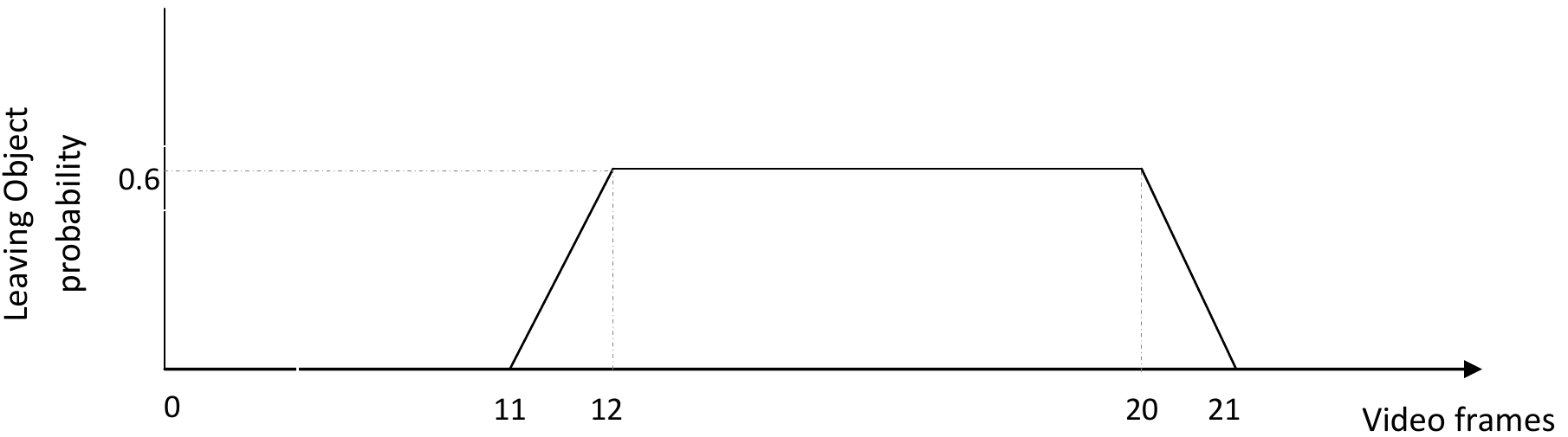}
\caption{Example of LTA probability fluctuation in the presence of a single initiation and a single termination.}
\label{fig:strongInitExample}
\end{figure}

In this section we illustrate the behaviour of Prob-EC in the case of fluents with a single initiation and a single termination. (Recall that there is no fixed relation between the number of initiations and terminations of a fluent.) Assume that \sarah\ is walking while simultaneously carrying a suitcase for 10 video frames. At frame 11, she leaves the suitcase on the floor and walks away from it. This causes the suitcase to `appear' in the low-level tracking system, triggering the \leave\ initiation condition expressed by rule \eqref{eq:leaving-inactive-init}. Suppose that this initiation condition has a probability of 0.6. At frame 20, \sarah\ picks up the suitcase, causing it to `disappear' and triggering termination condition expressed by rule \eqref{eq:leaving-exit-term}. Suppose that the termination condition also has a probability of 0.6.  Figure \ref{fig:strongInitExample} depicts the probability fluctuation of \leave. As can be seen in this figure, in the absence of any initiations after frame 11, 
the LTA persists entirely due to the law of inertia. The probability of this LTA is equal to the probability of the single initiation condition, that is, 0.6. Because this probability is above the chosen recognition threshold --- 0.5 in this example --- all frames taking place until \sarah\ picks up the suitcase will be counted as recognitions. If the probability of the initiation condition was below the threshold then \leave\ would not have been recognised.


Note that we may transform a fluent with a single initiation to a version of that fluent with multiple initiations --- consider the following formalisation:
\begin{align}
& \label{eq:leave-weak-init}
\begin{mysplit}
\initiatedAt( \mathit{leaving\_object\_mi(P,\ Obj)\val\true,\ T} ) \leftarrow \\
\quad \holdsAt(\mathit{leaving\_object(P,\ Obj)\val\true,\ T)} 
\end{mysplit}
\end{align}
$\mathit{leaving\_object\_mi}$ may have multiple initiations --- it is initiated as long as \linebreak\leave\ holds. While this transformation would have been beneficial for our experiments in the CAVIAR dataset, since we would be able to augment the probability of $\mathit{leaving\_object\_mi}$ through multiple initiations and eventually surpass the chosen recognition threshold, it introduces some subtle perils. Consider, for example, a scenario in which a \leave\ LTA is initiated with a small probability, such as 0.1, indicating that the sensor's confidence about the STA appearing in rule \eqref{eq:leaving-inactive-init} is low. Prob-EC will then compute that \leave\ holds with a probability of 0.1 up until the video frame, if any, where the object in question is picked up. These positives will be correctly discarded under most recognition thresholds given their small probabilities. $\mathit{leaving\_object\_mi}$, however, will continue to augment its probability, eventually surpassing the chosen threshold 
and thus producing a potentially large number of False Positives (FP). While it is true that such situations do not arise in CAVIAR, it is very likely that they take place in other activity recognition applications.

\leave\ has a single termination. In this example, the probability of the termination condition drops the probability of \leave\ below the chosen threshold of 0.5. In other examples, the absence of subsequent terminations may not allow the probability of a LTA to drop below the chosen recognition threshold, thus possibly resulting in false persistence.


\section{Experimental Evaluation}\label{sec:LTAR_EC_results}

\subsection{CAVIAR without Artificial Noise}\label{sec:exp-caviar-no-noise}

\begin{table}[t]
\begin{center}
\renewcommand{\arraystretch}{0.9}
\setlength\tabcolsep{3pt}
\begin{tabular}{ccccccc}
\hline\noalign{\smallskip}
\multicolumn{1}{c}{\textbf{LTA}} & \multicolumn{1}{c}{\textbf{TP}} &
\multicolumn{1}{c}{\textbf{FP}} & \multicolumn{1}{c}{\textbf{FN}} & 
\multicolumn{1}{c}{\textbf{Precision}} & \multicolumn{1}{c}{\textbf{Recall}} &
\multicolumn{1}{c}{\textbf{F-measure}}
\\
\noalign{\smallskip}
\hline
\noalign{\smallskip}
meeting & 3099 & 1910 & 525 & 0.619  & 0.855 & 0.718  \\[2pt]

moving & 4008 & 2162 & 2264 & 0.650 & 0.639 &  0.644  \\[2pt]

fighting & 531 & 97 & 729 & 0.421 & 0.845 & 0.562   \\[2pt]

leaving object & 143 & 1539 & 55 & 0.085 & 0.722 & 0.152  \\[2pt]
\hline
\end{tabular}
\end{center}
\caption{ True Positives (TP), False Positives (FP), False Negatives (FN), Precision, Recall and F-measure on CAVIAR without artificial noise.}\label{tbl:stats}
\end{table}

Our empirical analysis is based on the 28 surveillance videos of the CAVIAR dataset which contain, in total, 26419 video frames. These frames have been manually annotated by the CAVIAR team to provide the ground truth for STA and LTA (we performed very minor editing of the annotation in order to introduce a STA for abrupt motion). According to the manual annotation of the dataset, all STA are associated with a probability of 1. Table \ref{tbl:stats} shows the recognition results in terms of True Positives (TP), False Positives (FP), False Negatives (FN), Precision, Recall and F-measure. These results have been produced by computing queries of the form \holdsAt$\mathit{(LTA\val\true,\ T)}$. Given that STA have probability of 1, Prob-EC has identical results to Crisp-EC. 

Particularly notable in the results is the low Precision for the \leave\ LTA, owing to a substantial number of FP. This is due to the problematic annotation of CAVIAR with respect to this LTA. For example, in video 14, the object that is left at frame 946, triggering an initiation of \leave, is picked up (`disappears') at frame 1354. However, the relevant annotation mistakenly reports
that \leave\ stops occurring at frame 996. We therefore end up with 358 FP which could have been avoided with a more consistent annotation of this video. Similarly, in the annotation of videos 17 and 18, a large number of annotated frames are missing. Video 16 includes a particularly interesting case of \leave. In this video, a person leaves a bag next to a chair, exits the scene, re-enters after a couple of seconds and picks up the bag. When the person re-enters, he is assigned a new identifier (this is common in CAVIAR). Various complications arise due to this. The original \leave\ activity is not terminated by our rules when the person in question `disappears'. This is deliberate on our part: we choose to terminate \leave\ when the object is picked up rather than when the person that leaves it `disappears' from the sensor's point  of view. We thus emphasize on time-points in which a package might be unattended. The CAVIAR annotation, however, views the \leave\ LTA from a different perspective and thus 
assumes that the activity is terminated when the person `disappears'. This difference in perspective leaves us with a substantial number of FP, one for each frame that the person is not present in the scene. When the person re-enters the scene, CAVIAR provides the person with a new identifier and resumes the annotation of \leave\ with a \emph{$< $new\_person\_id, same\_object\_id$ >$} tuple. After a couple of frames, the person (described by \emph{new\_person\_id}) picks up the object, terminating a $\mathit{leaving\_object(new\_person\_id, same\_object\_id)}$ LTA occurrence which Crisp-EC never initiated in the first place. Thus, in addition to a substantial number of FP, we also generate 55 FN (see Table \ref{tbl:stats}) because Crisp-EC never recognises the new \leave\ activity.

\subsection{CAVIAR with Artificial Noise}

As mentioned above, according to the manual annotation of the CAVIAR dataset all STA are associated with a probability of 1. In real-world activity recognition applications it is unrealistic to assume that STA will be detected with certainty. In order to experiment with a more realistic setting, we added artificial noise to the dataset, in the form of probabilities attached to the input facts --- STA and related coordinate and orientation information. 
Our experimental procedure may be summarised as follows. 

\begin{itemize}
\item We inject noise into CAVIAR:
  \begin{enumerate}
  \item We add probabilities to STA. Toward this end, we use a Gamma distribution with a varying mean in order to represent different levels of noise.
  \item In addition to STA, we add probabilities to their associated coordinate and orientation fluents. Although we use the same Gamma distribution for this step, STA are not required to have the same probability as their associated coordinate and orientation fluents.
  \item On top of the above, we introduce spurious STA, that is, STA that are not part of the original CAVIAR dataset. We augment the frames in which there is a \walking\ STA with another \walking\ STA and related coordinate and orientation information about an entity that does not exist in CAVIAR. We use a uniform distribution to choose the frames that will be augmented with spurious facts. The probability of a spurious fact at frame $t$ is $1{-}p$ where $p$ is the probability of some CAVIAR \walking\ STA at $t$ (recall that $p$ was computed by the Gamma distribution).
 \end{enumerate}
 Thus we end up with three different noisy versions of CAVIAR. To facilitate the presentation that follows, we will call the three aforementioned approaches to noise injection `smooth', `intermediate' and `strong' noise.

\item We feed these data to Prob-EC and filter its output --- which is a series of positives of the form  $\mathit{Prob::\holdsAt(LTA \val \true,\ T)}$ --- to keep only the positives with probability above a chosen threshold, indicating that we trust these positives to be accurate. 

\item We filter each noisy version of the dataset by erasing the facts with probability below the chosen threshold. We retain the facts with probability above the threshold, removing their probability values. Thus we assume that such facts have been tracked with certainty. For `smooth' noise, this step means that we remove a certain amount of the CAVIAR STA, while for `intermediate' noise we additionally remove coordinate and orientation fluents. Furthermore, for `strong' noise we keep a certain amount of spurious information. 

\item We give these filtered versions of CAVIAR as input to Crisp-EC. With this step we aim to estimate the impact of environmental noise on Crisp-EC, by assuming that Crisp-EC can only reason on facts that have been tracked with relative certainty, that is, facts with probability above the chosen threshold.

\item We compare the performance of Crisp-EC and Prob-EC.
\end{itemize}

We repeated the above experimental procedure 16 times, once for each Gamma distribution mean value between 0.5 and 8.0 inclusive, with a step of 0.5. Noise is added randomly; for example, a STA that is erased from Crisp-EC's input for Gamma mean 6.0 might be present at the dataset produced for Gamma mean 6.5.  In our implementation, the higher the mean, the lower the probabilities attached to the CAVIAR input facts and the higher the probability of the spurious facts, indicating a higher amount of noise. 

\begin{figure}[t]
        \centering
        \begin{subfigure}[b]{.65\textwidth}
                \centering
                \includegraphics[width=\textwidth]{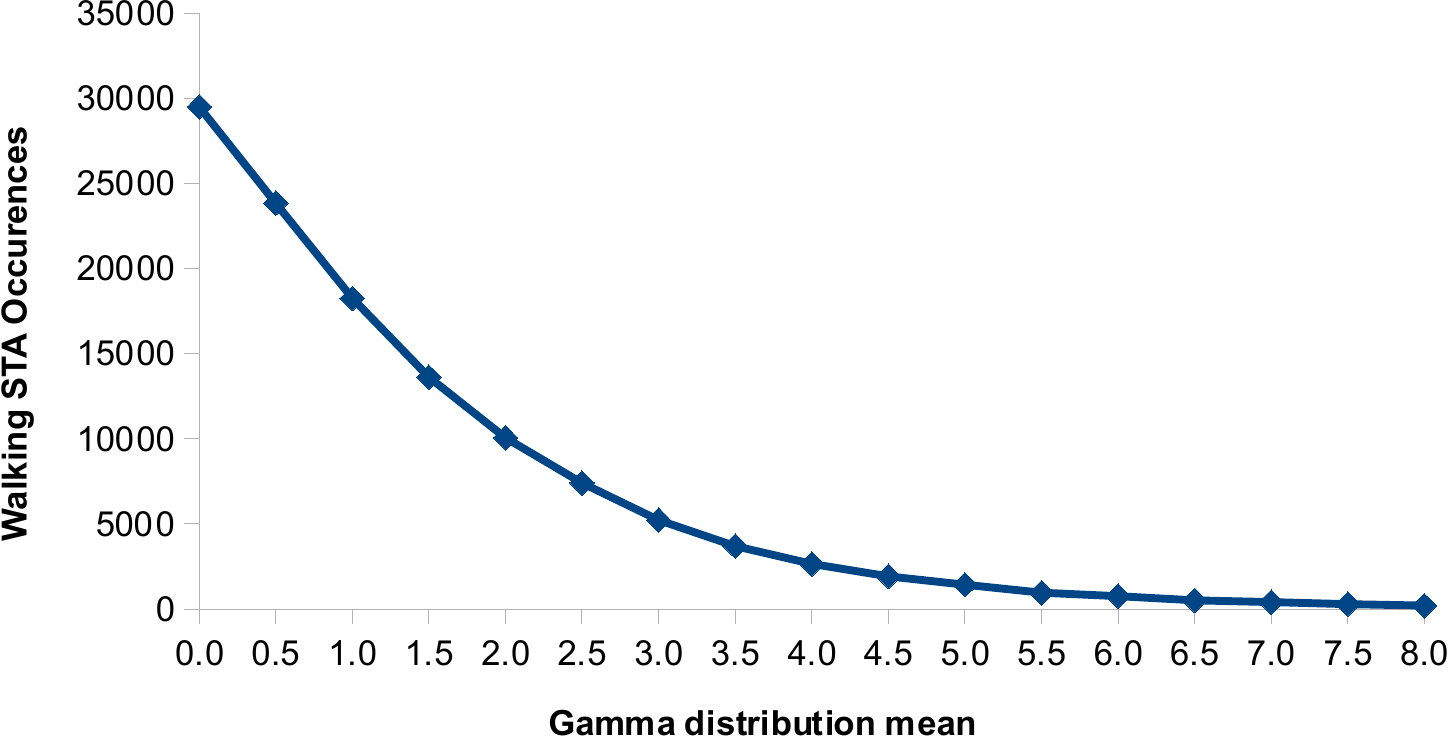}
                \caption{(a) CAVIAR STA.}
                \label{fig:STAOcc}
        \end{subfigure}\linebreak\linebreak 
        ~ 
        \begin{subfigure}[b]{.65\textwidth}
                \centering
                \includegraphics[width=\textwidth]{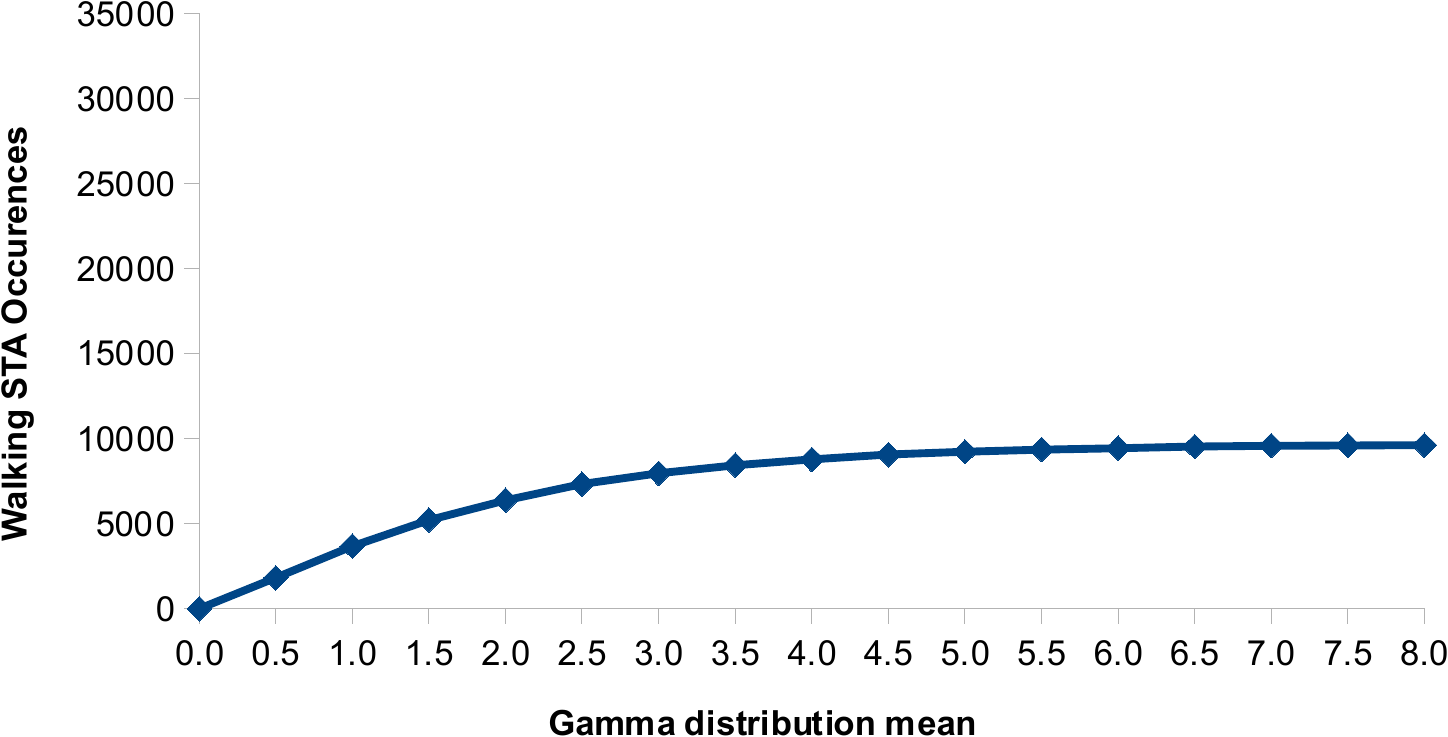}
                \caption{(b) Spurious STA.}
                \label{fig:GhostSTAOcc}
        \end{subfigure}
        \caption{Occurrences of \walking\ STA with probability above 0.5 per Gamma mean value. Mean value 0.0 represents the CAVIAR dataset without artificial noise.}\label{fig:occ}
\end{figure}

Figure \ref{fig:occ}(a) shows an aspect of noise injection. In this figure the number of occurrences of the \walking\ STA with probability above the 0.5 threshold is plotted. Note that the numbers of \walking\ occurrences shown in this figure do not include the spurious facts introduced in the `strong' noise experiments. The amount of \walking\ occurrences drops exponentially as we increase the level of noise, that is, the Gamma mean value. All other STA follow the same pattern. In the case of `intermediate noise', the occurrences of coordinate and orientation fluents also drop exponentially. 

Figure \ref{fig:occ}(b) shows the number of occurrences of the spurious \walking\ STA --- these are introduced in the `strong' noise experiments --- with probability above the 0.5 threshold. The amount of spurious facts increases exponentially as we increase the level of noise (Gamma mean value).


Noise injection may significantly reduce recognition accuracy. The sections that follow illustrate this. However, given that there are mistakes in the original dataset (that is, CAVIAR without artificial noise), it may occur that a FP in the original dataset becomes a True Negative (TN) in a noise-altered version of CAVIAR. Similarly for FN. To demonstrate this with an example, assume that two people are \move\ along together on a pavement. Suddenly they have to step aside to allow a handicapped person full access to the pavement. This would fire the `walk away' termination condition (see rule \eqref{eq:moving-term}) because the distance between the two people would exceed the pre-specified threshold. This does not mean that the people have stopped \move\ --- they merely had to distance themselves momentarily. 
Firing --- erroneously --- the `walk away' termination condition creates FN for some video frames.
If, however, the \walking\ STA gets a very low probability during the frames at which the people make way to the handicapped person, \move\ will not be terminated, which, in this special case, adds TP to the evaluation. 

\begin{figure}[h]
        \centering
        \begin{subfigure}[b]{.5\textwidth}
                \centering
                \includegraphics[width=\textwidth]{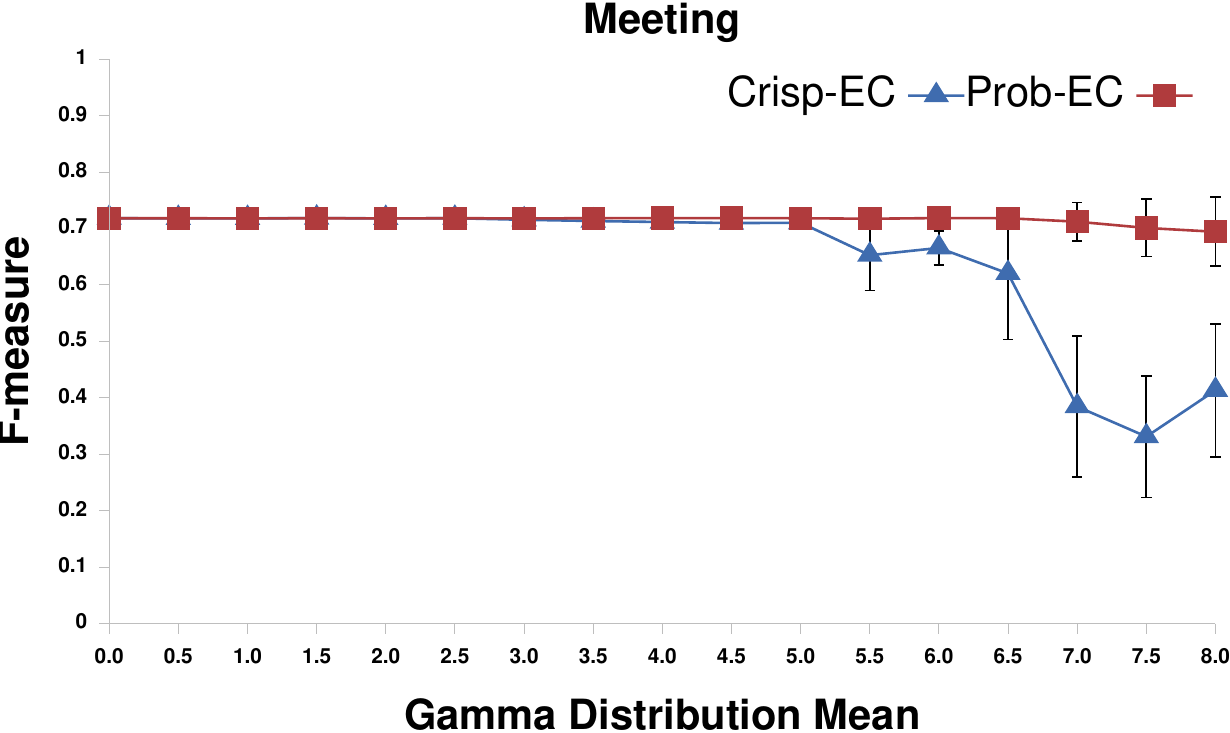}
                \caption{(a)}
                \label{fig:exp-original-smooth-fmeasure-05-meeting}
        \end{subfigure}%
        ~ 
        \begin{subfigure}[b]{.5\textwidth}
                \centering
                \includegraphics[width=\textwidth]{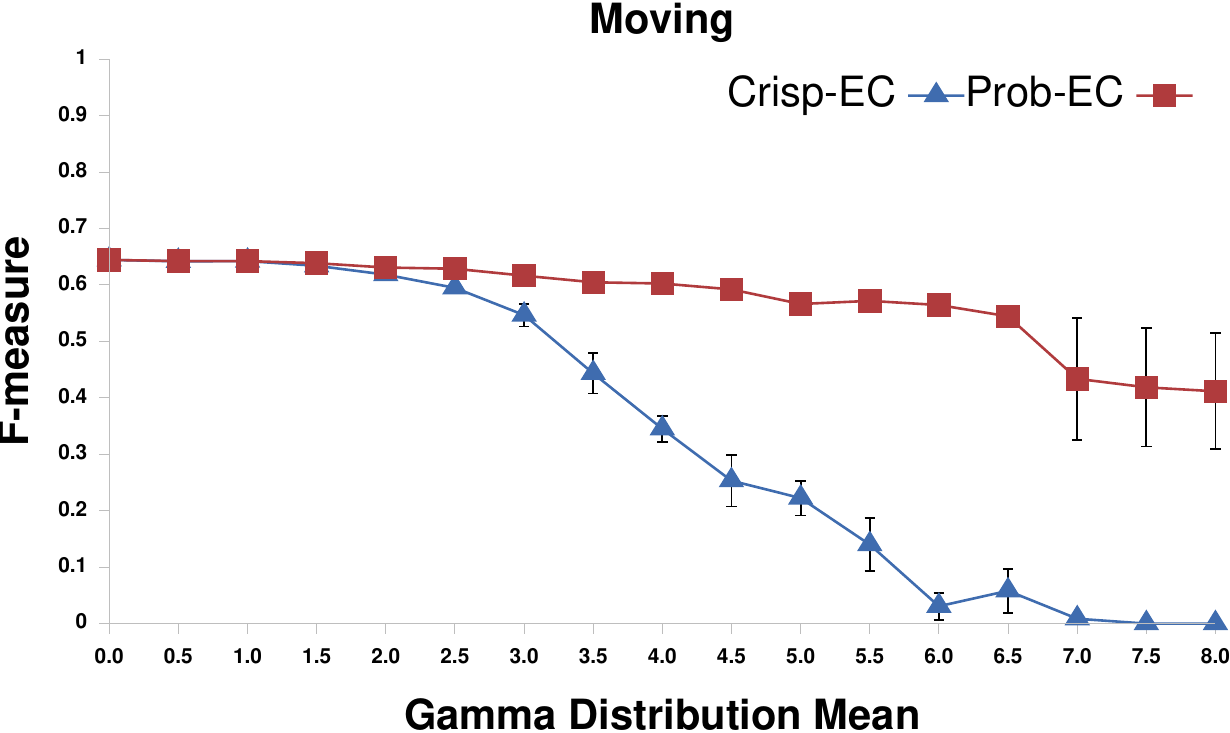}
                \caption{(b)}
                \label{fig:exp-original-smooth-fmeasure-05-moving}
        \end{subfigure}
        \linebreak\linebreak 
        \begin{subfigure}[b]{.5\textwidth}
                \centering
                \includegraphics[width=\textwidth]{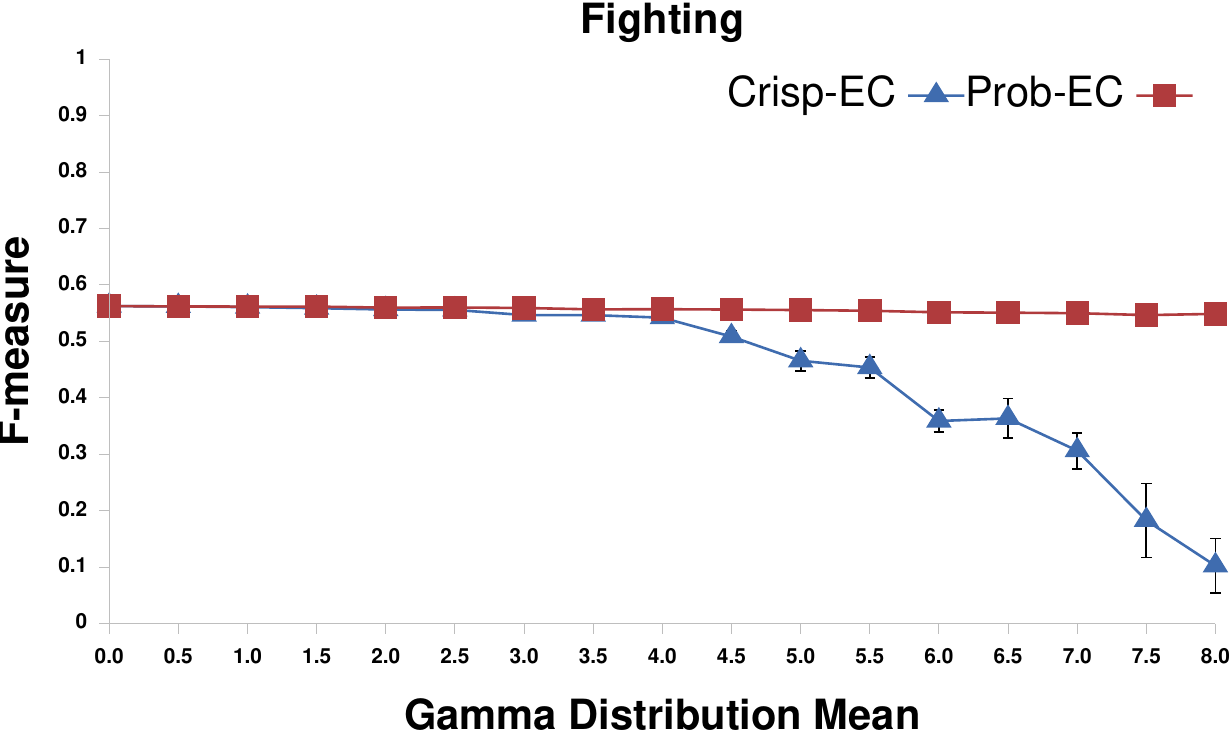}
                \caption{(c)}
                \label{fig:exp-original-smooth-fmeasure-05-fighting}
        \end{subfigure}~
	\begin{subfigure}[b]{.5\textwidth}
                \centering
                \includegraphics[width=\textwidth]{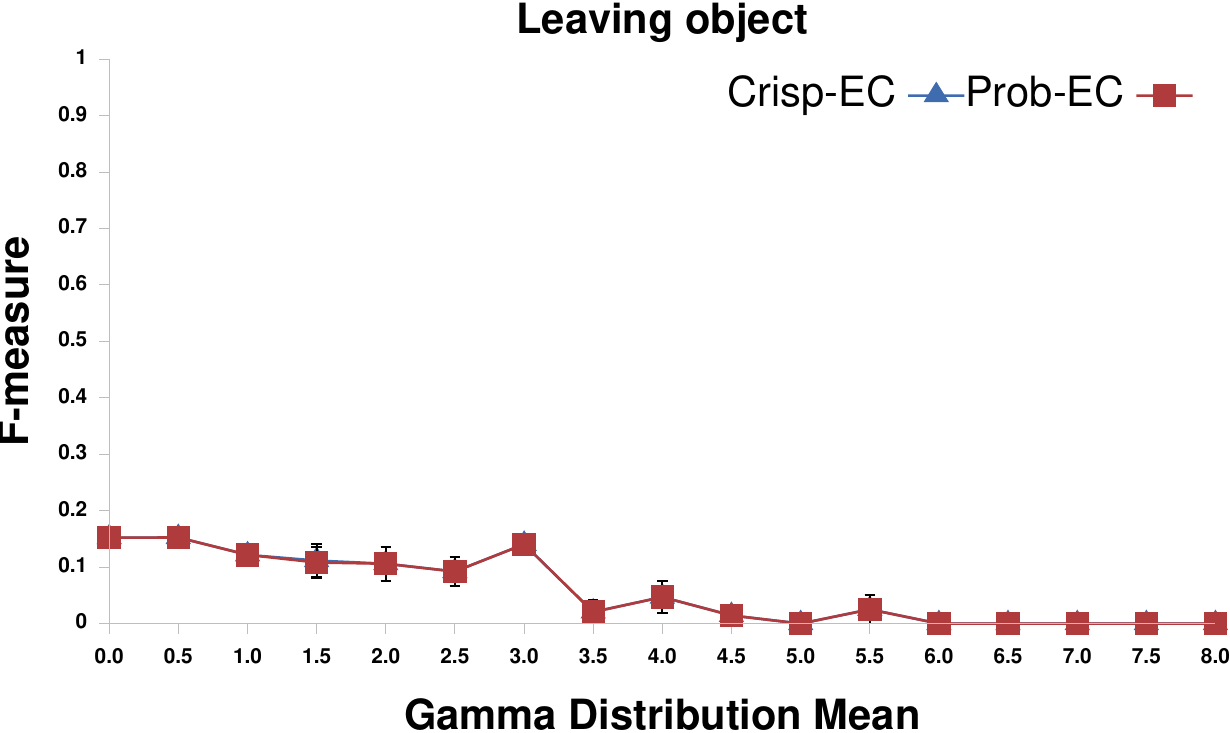}
                \caption{(d)}
                \label{fig:exp-original-smooth-fmeasure-05-leaving-object}
        \end{subfigure}
        \caption{Crisp-EC and Prob-EC F-measure per Gamma mean value under `smooth' noise and a 0.5 threshold.}\label{fig:exp-original-smooth-fmeasure-05}
\end{figure}

\subsubsection{`Smooth' Noise Experiments}\label{subsec:smoothNoiseExps}

Figure \ref{fig:exp-original-smooth-fmeasure-05} compares the recognition accuracy of Crisp-EC and Prob-EC  in terms of F-measure under `smooth' noise using a 0.5 threshold. In all experiments we plot the F-measure per Gamma distribution mean averaged over 5 different runs, that is, each point in the diagram is the average of 5 different F-measure values. The vertical error bars display the standard deviations. 

The recognition accuracy of Prob-EC is higher than that of Crisp-EC in \meet, \move\ and \fight. In \meet\ (see Figure \ref{fig:exp-original-smooth-fmeasure-05}(a)), the accuracy of Crisp-EC starts falling from Gamma mean 5.5 onwards, because the \activeb\ and \inactive\ STA tend to be erased from its input, since they receive probability below 0.5 --- the chosen threshold in the experiments presented in Figure \ref{fig:exp-original-smooth-fmeasure-05} --- when we generate noise. Prob-EC, on the other hand, is able to initiate \meet\ with a certain degree of probability (recall that rules \eqref{eq:meet-inactive-init} and \eqref{eq:meet-active-init} express the conditions in which \meet\ is initiated). After a number of frames, the repeated initiation of \meet\ leads to a \holdsAt\ probability of 0.5 or above, and thus Prob-EC eventually recognises \meet\ (refer to Section \ref{sec:multiple-initiations} for a detailed explanation of this behaviour). Those initiation conditions occur very frequently in CAVIAR.
 Therefore, \meet\ ends up being recognised with a probability close to 1.

In the case of \move\ (see Figure \ref{fig:exp-original-smooth-fmeasure-05}(b)), it is the loss of multiple occurrences of \walking\ (due to `smooth' noise) that causes Crisp-EC to suffer from numerous FN (see rule \eqref{eq:moving-init} for the initiation condition of \move). Prob-EC, on the other hand, uses repeated initiation to eventually surpass the 0.5 recognition threshold. The accuracy of Crisp-EC in \move\ deteriorates earlier than that of \meet. This occurs because the initiation condition of \move\ bears two probabilistic conjuncts in its body, corresponding to two separate occurrences of the \walking\ STA. Therefore, it is affected by noise twice. The initiation conditions of \meet\ may sometimes have two probabilistic conjuncts, but usually they have just one --- the \negate$_1$ conditions are usually crisp. 

Similar to \move\ and \meet, Prob-EC fares better than Crisp-EC in \fight\ (see Figure \ref{fig:exp-original-smooth-fmeasure-05}(c)). For high levels of noise, that is, high Gamma mean values, Crisp-EC has trouble initiating \fight\ (see rule \eqref{eq:fighting-init-abrupt}), whereas Prob-EC uses repeated initiation to surpass the 0.5 recognition threshold. Prob-EC manages to surpass this threshold despite the relatively short duration of fights, compared to other LTA.

\leave\ is an interesting special case, owing to the fact that this LTA is recognised through a single initiation (see Section \ref{sec:single-initiation}). In the CAVIAR videos, a person leaves an object after a few frames in which he is walking. 
In rare situations it is possible that either Crisp-EC or Prob-EC miss the recognition of \person\ due to noise and, consequently, the recognition of \leave. Such cases did not arise in our `smooth' noise experiments.
In these experiments, Prob-EC and Crisp-EC are equally accurate (see Figure \ref{fig:exp-original-smooth-fmeasure-05}(d)). When \person\ is recognised by Crisp-EC, it is recognised with a sufficiently high probability by Prob-EC and vice versa. 

LTA in CAVIAR are usually terminated when an entity `disappears' or when people walk away from each other. In the latter case, the probability of the LTA termination depends solely on that of \walking\ (see, for example, rule \eqref{eq:moving-term}). 
In general, probabilistic terminations are similar to probabilistic initiations. 
When a termination condition is repeatedly fired with probabilities below 0.5 then Prob-EC eventually stops recognising the LTA, whereas Crisp-EC does not, producing FP.

\begin{figure}[H]
        \centering
        \begin{subfigure}[b]{.5\textwidth}
                \centering
                \includegraphics[width=\textwidth]{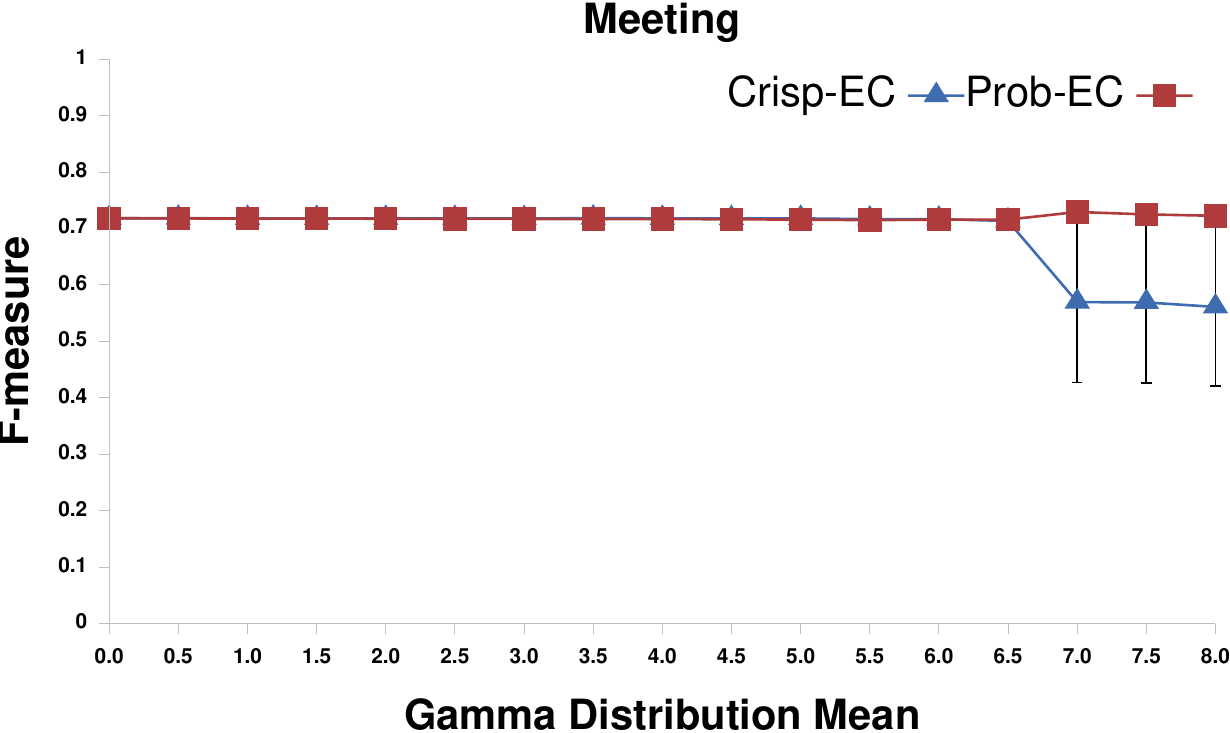}
                \caption{Threshold$\val$0.3}
        \end{subfigure}%
        ~ 
        \begin{subfigure}[b]{.5\textwidth}
                \centering
                \includegraphics[width=\textwidth]{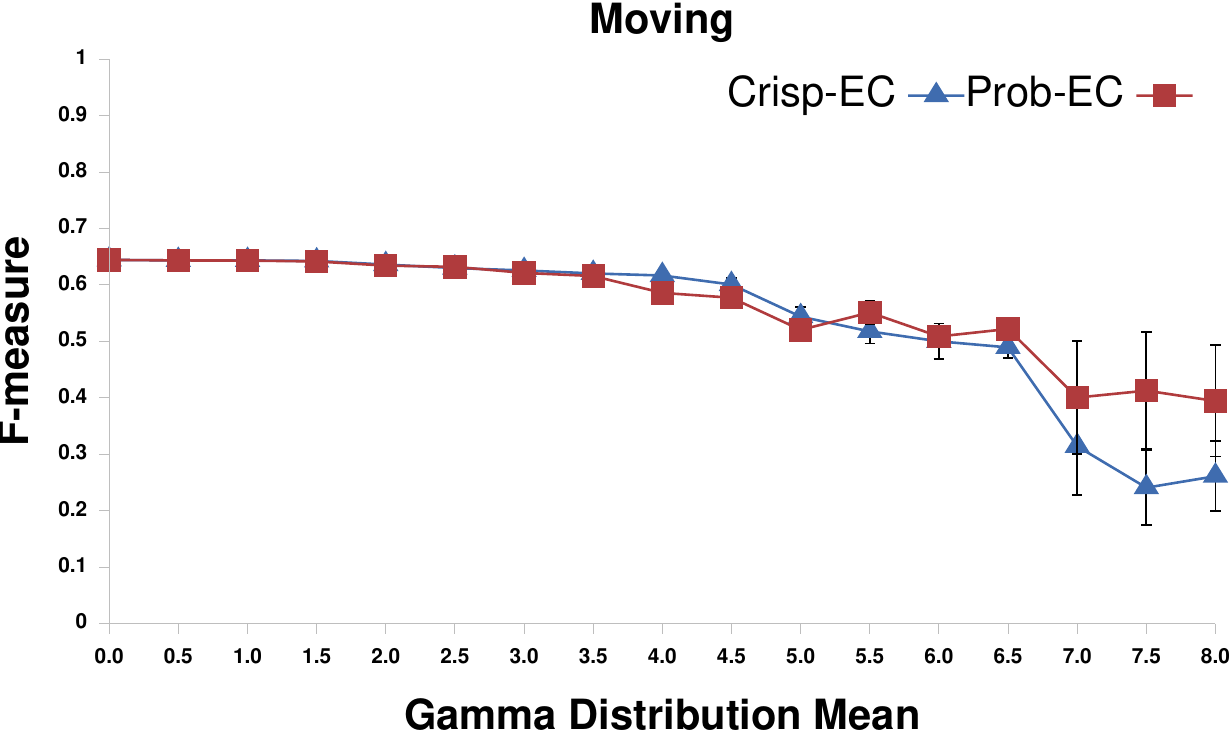}
                \caption{Threshold$\val$0.3}
        \end{subfigure}
        \linebreak\linebreak 
        \begin{subfigure}[b]{.5\textwidth}
                \centering
                \includegraphics[width=\textwidth]{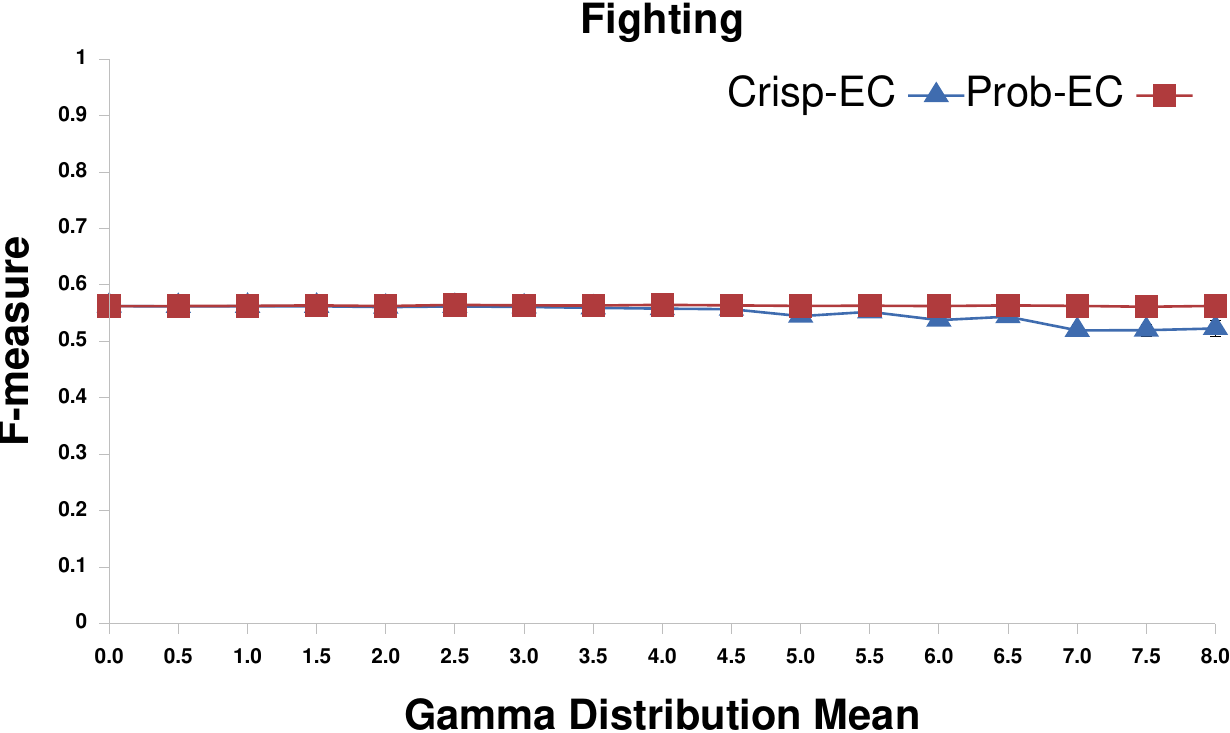}
                \caption{Threshold$\val$0.3}
        \end{subfigure}~
	\begin{subfigure}[b]{.5\textwidth}
                \centering
                \includegraphics[width=\textwidth]{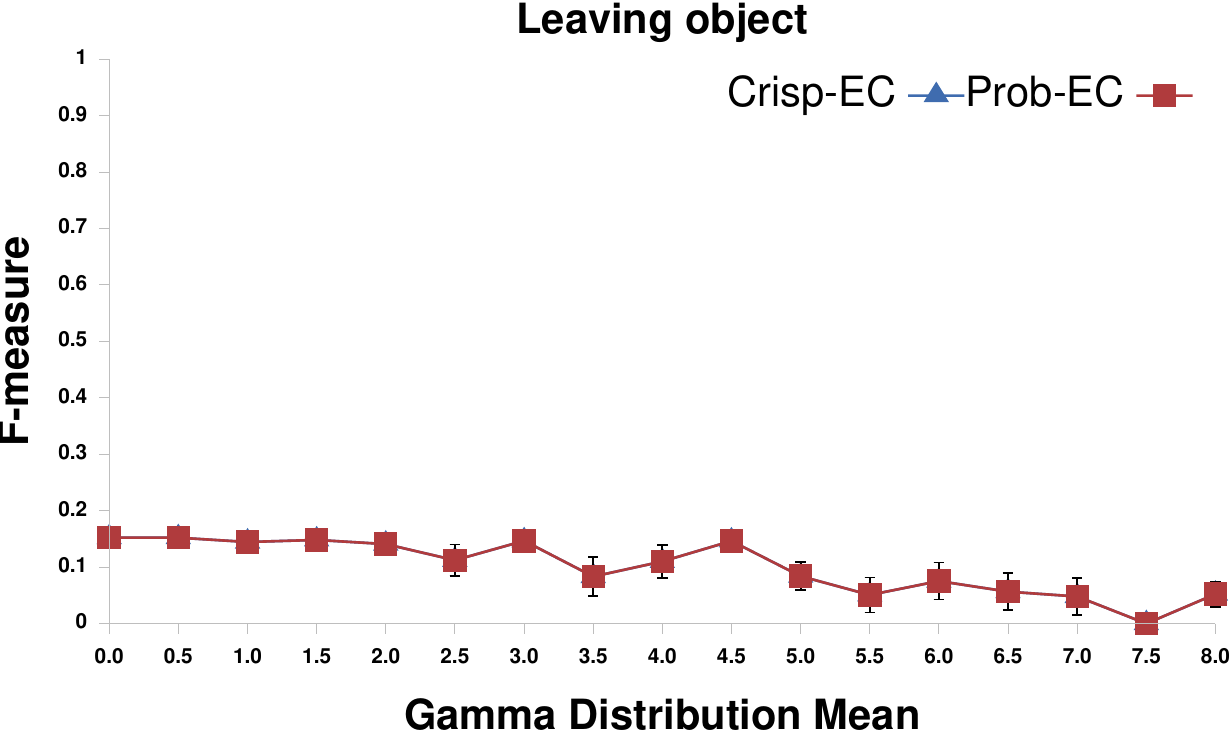}
                \caption{Threshold$\val$0.3}
        \end{subfigure}
	\linebreak\linebreak
        \begin{subfigure}[b]{.5\textwidth}
                \centering
                \includegraphics[width=\textwidth]{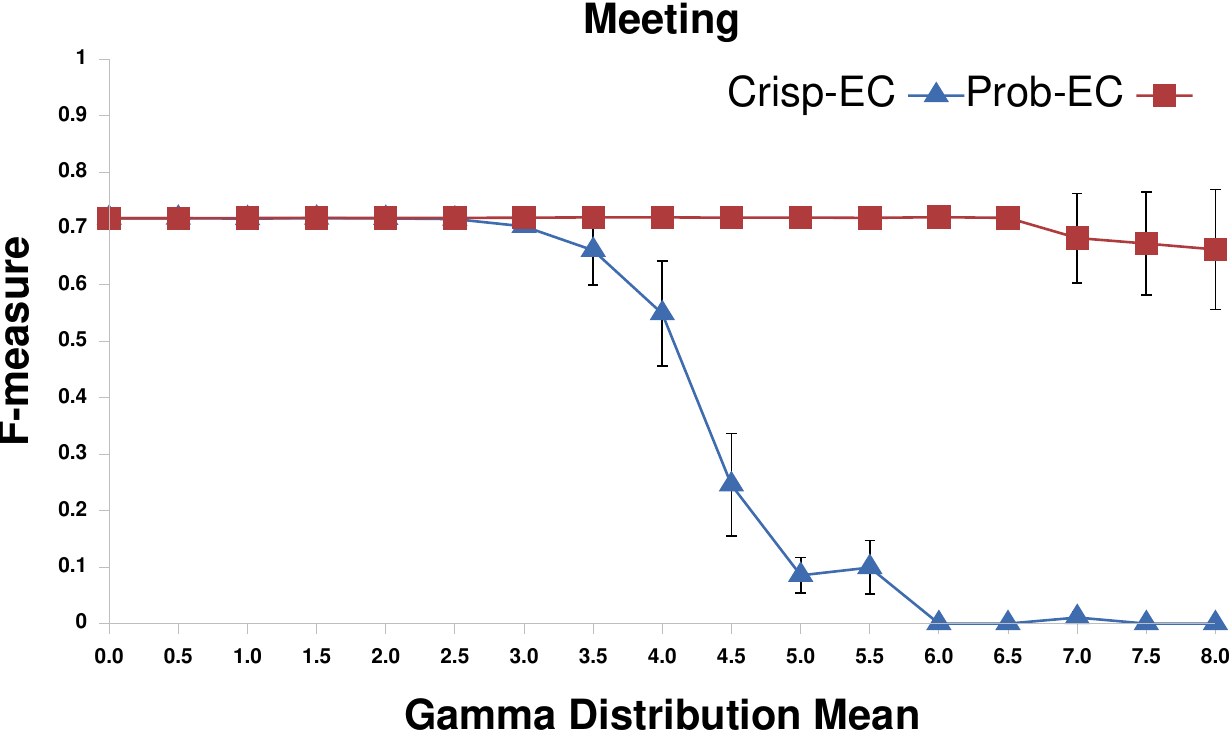}
                \caption{Threshold$\val$0.7}
        \end{subfigure}%
        ~ 
        \begin{subfigure}[b]{.5\textwidth}
                \centering
                \includegraphics[width=\textwidth]{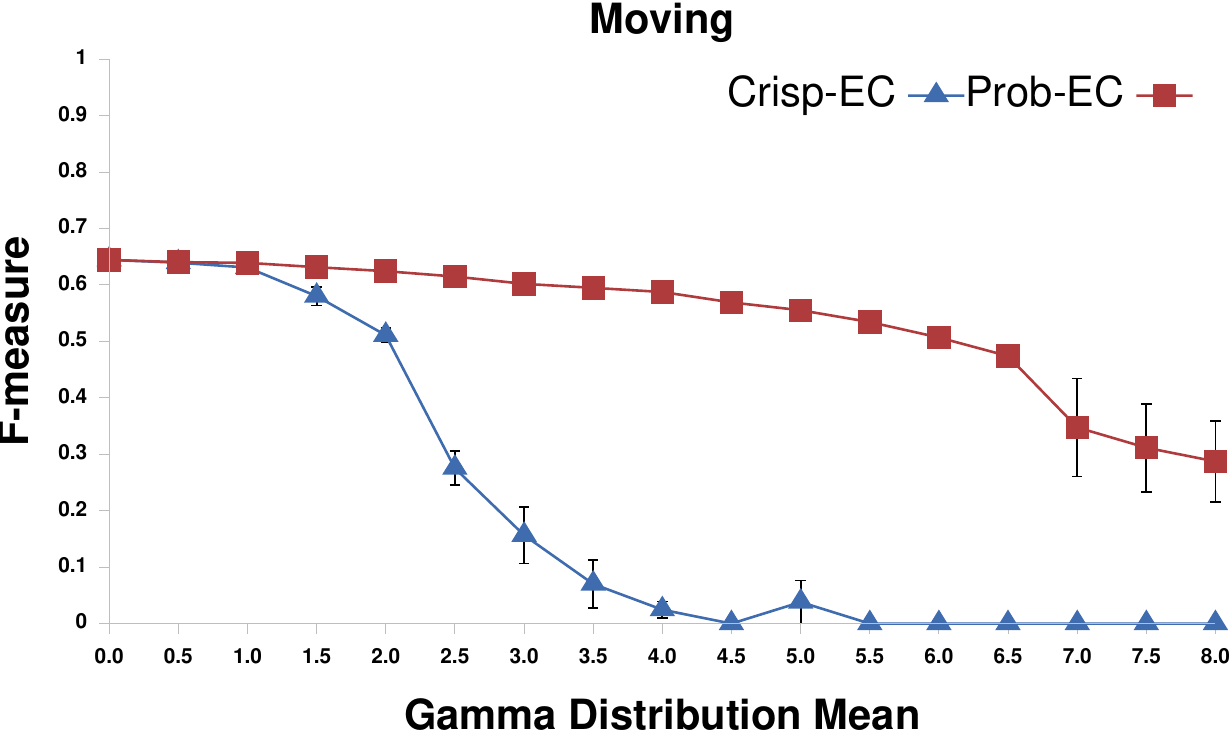}
                \caption{Threshold$\val$0.7}
        \end{subfigure}
        \linebreak\linebreak 
        \begin{subfigure}[b]{.5\textwidth}
                \centering
                \includegraphics[width=\textwidth]{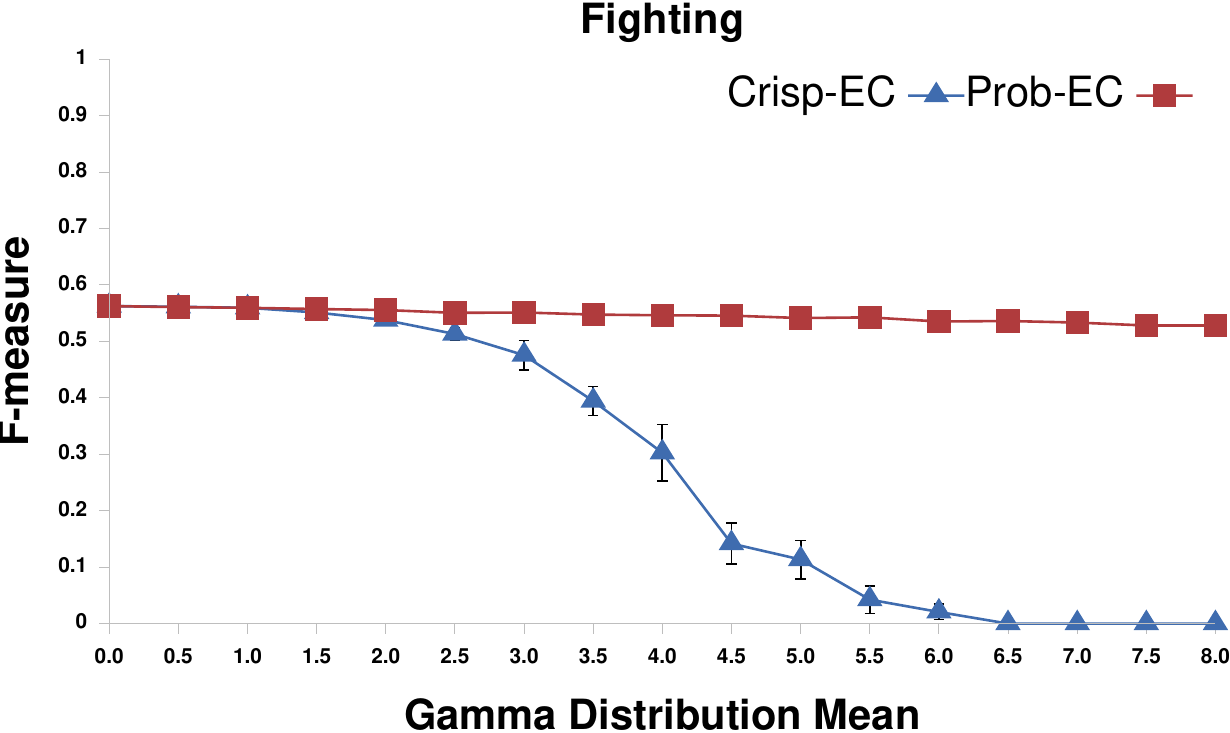}
                \caption{Threshold$\val$0.7}
        \end{subfigure}~
	\begin{subfigure}[b]{.5\textwidth}
                \centering
                \includegraphics[width=\textwidth]{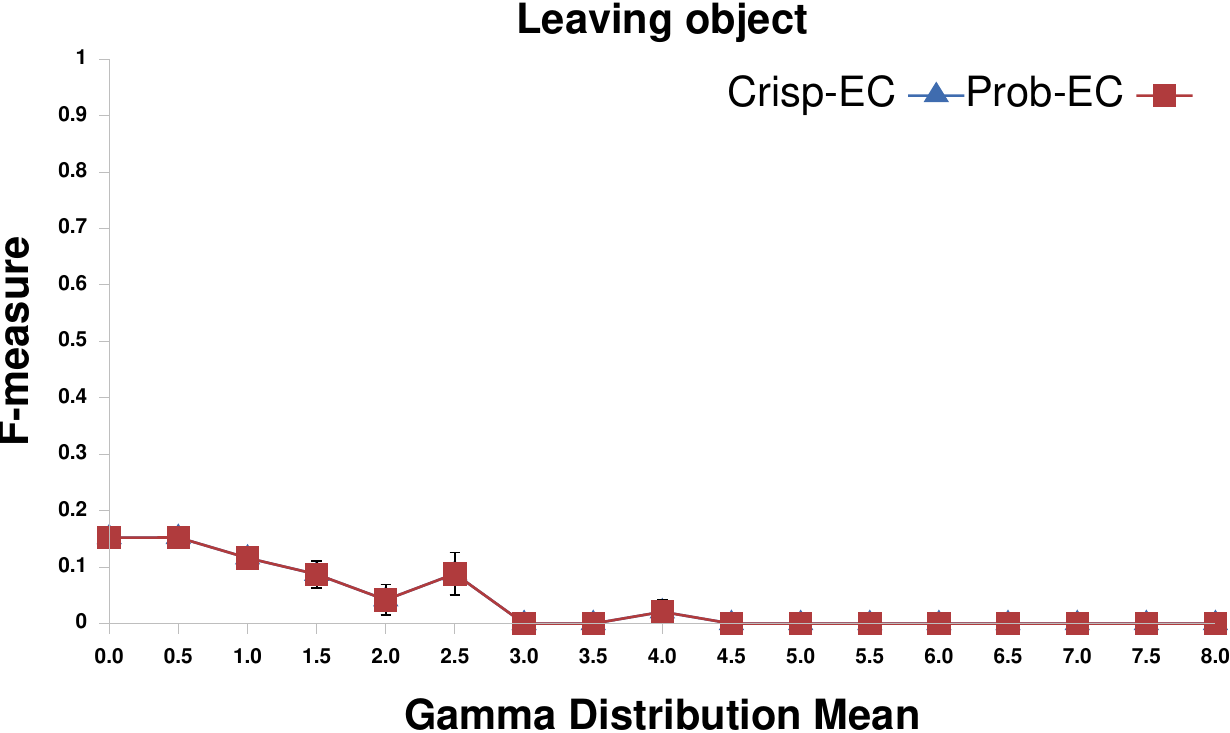}
                \caption{Threshold$\val$0.7}
        \end{subfigure}
        \caption{Crisp-EC and Prob-EC F-measure per Gamma mean value under `smooth' noise. In the first four figures the threshold is 0.3 while in the last four figures the threshold is 0.7.}\label{fig:exp-original-smooth-fmeasure-03-07}
\end{figure}

Figure \ref{fig:exp-original-smooth-fmeasure-03-07} compares the recognition accuracy of Crisp-EC and Prob-EC using 0.3 and 0.7 thresholds. This figure shows that Prob-EC is only slightly affected by the threshold change. This is due to the fact that the probabilities of most positives in Prob-EC are either very small ($<$0.3) or very high ($>$0.7). 
On the other hand, the recognition accuracy of Crisp-EC is highly affected by the choice of threshold: the lower the threshold the better the accuracy of Crisp-EC. This is expected given the nature of the `smooth' noise experiments. As the threshold decreases, the input of Crisp-EC gets closer to the original, artificial noise-free dataset.

\begin{figure}[h]
        \centering
        \begin{subfigure}[b]{.5\textwidth}
                \centering
                \includegraphics[width=\textwidth]{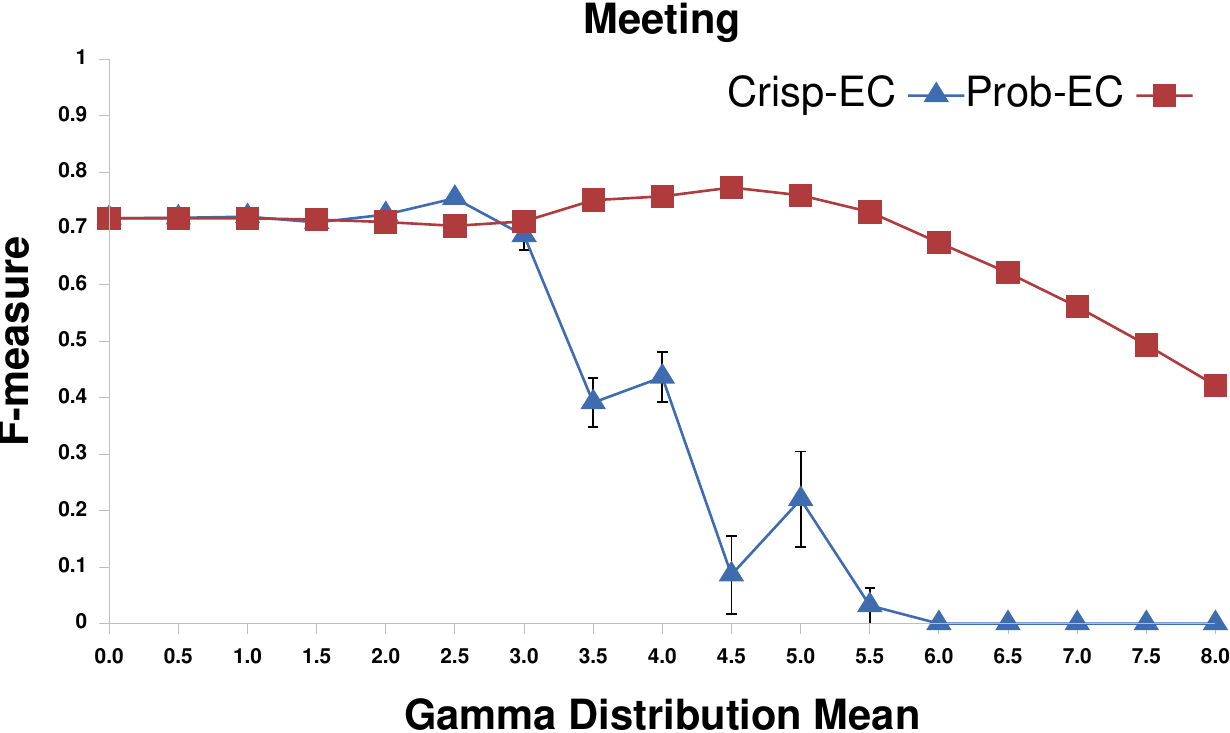}
                \caption{(a)}
                \label{fig:exp-original-intermediate-fmeasure-05-meeting}
        \end{subfigure}%
        ~ 
        \begin{subfigure}[b]{.5\textwidth}
                \centering
                \includegraphics[width=\textwidth]{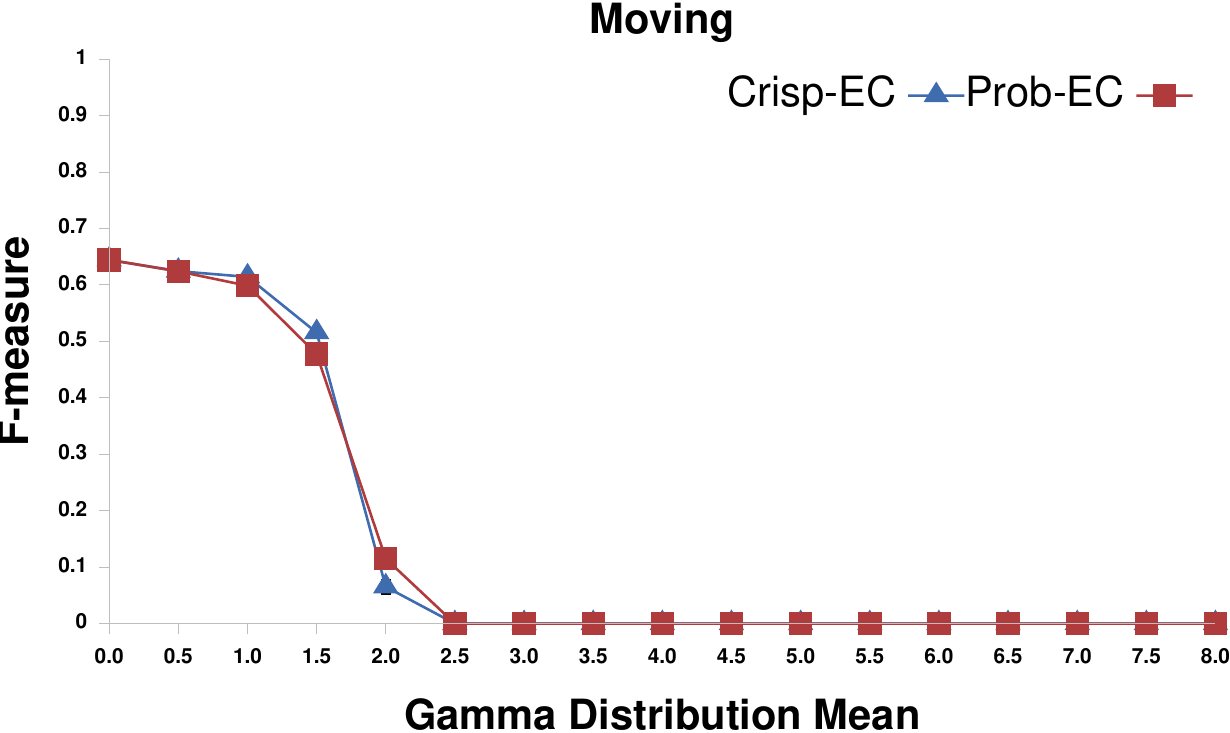}
                \caption{(b)}
                \label{fig:exp-original-intermediate-fmeasure-05-moving}
        \end{subfigure}
        \linebreak\linebreak 
        \begin{subfigure}[b]{.5\textwidth}
                \centering
                \includegraphics[width=\textwidth]{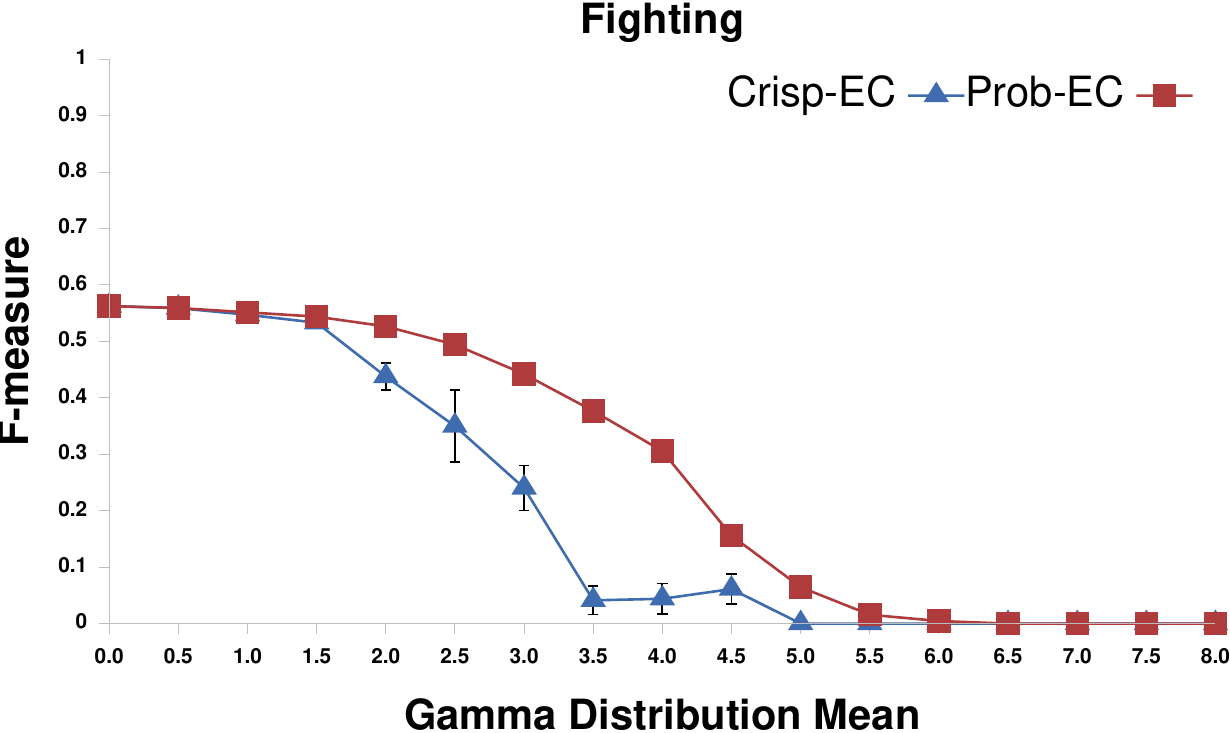}
                \caption{(c)}
                \label{fig:exp-original-intermediate-fmeasure-05-fighting}
        \end{subfigure}~
	\begin{subfigure}[b]{.5\textwidth}
                \centering
                \includegraphics[width=\textwidth]{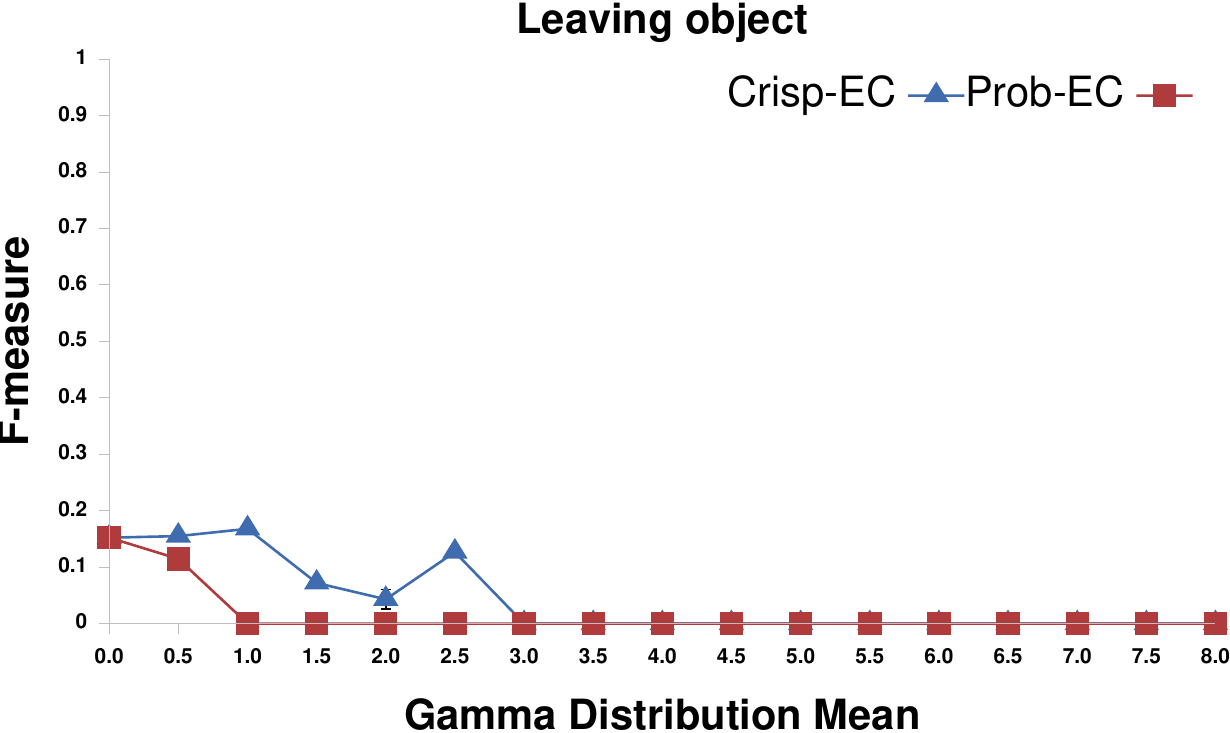}
                \caption{(d)}
                \label{fig:exp-original-intermediate-fmeasure-05-leaving-object}
        \end{subfigure}
        \caption{Crisp-EC and Prob-EC F-measure per Gamma mean value under `intermediate' noise and a 0.5 threshold.}\label{fig:exp-original-intermediate-fmeasure-05}
\end{figure}

\subsubsection{`Intermediate' Noise Experiments}\label{subsec:intermediateNoiseExps}

Figure \ref{fig:exp-original-intermediate-fmeasure-05} compares the recognition accuracy of Crisp-EC and Prob-EC  under `intermediate' noise using a 0.5 threshold. 
Overall, attaching probabilities to the coordinates and orientation, in addition to STA, has reduced the accuracy of both Crisp-EC and Prob-EC. Compared to `smooth' noise, the difference between Prob-EC and Crisp-EC in \meet\ is bigger (see Figure \ref{fig:exp-original-intermediate-fmeasure-05}(a)). This occurs because, in addition to losing the \activeb\ and \inactive\ STA due to noise, Crisp-EC now has trouble proving that entities are \close, because under `intermediate' noise we also remove coordinate-related information, required to compute the distance between two entities. Thus, even in cases where the \activeb\ or \inactive\ STA are present and indicate that the relevant frame might be an initiation condition for \meet, Crisp-EC is unable to prove that the two entities are \close\ enough to initiate the LTA. Prob-EC, on the other hand, uses repeated initiation to eventually break the 0.5 barrier and produce recognitions.


Concerning \move\ (see Figure \ref{fig:exp-original-intermediate-fmeasure-05}(b)), Crisp-EC suffers again from the loss of the \walking\ STA. Under `intermediate noise', this conclusion is more striking: after Gamma mean 2.5, Crisp-EC is unable to produce a single positive. This is because, in addition to the \walking\ STA and associated coordinate fluents being erased from Crisp-EC's input, orientation fluents are also candidates for removal. As a result, even in cases where both entities potentially involved in a \move\ activity are believed to be \walking\ close to each other by the low-level tracking system, the absence of orientation information leads to an inability to initiate \move.


What is perhaps more interesting is that Prob-EC performs as bad as Crisp-EC in \move\ (see Figure \ref{fig:exp-original-intermediate-fmeasure-05}(b)). This is because whenever faced with an initiation condition for \move, Prob-EC has to calculate the probability of this initiation condition as a product of 6 probabilities in total. (Recall from rule \eqref{eq:close} that \close\ is defined in terms of two coordinate input facts. It therefore contributes as the product of 2 probabilities.) Consequently, Prob-EC has trouble surpassing the 0.5 recognition threshold we require to trust its positives. Even in cases of near-certainty about some conditions of the input, ProbLog's independence assumption leads to very low probability values. Consider, for example, two entities, $\mathit{id_1}$ and $\mathit{id_2}$, whose STA (in our case, \walking) and associated information (coordinates, orientation) are all tracked with a probability of 0.8. Whereas Crisp-EC will be able to initiate \move, since all the facts will 
make their appearance in the input, Prob-EC will produce an initiation condition probability of (0.8)$^6\val$0.262. Consequently, we will not trust the probability of the relevant \holdsAt\ query. 
Furthermore, due to the fact that each initiation condition has usually a very low probability, repeated initiation does not overcome the 0.5 threshold. 
%

With respect to \fight\ (see Figure \ref{fig:exp-original-intermediate-fmeasure-05}(c)), Prob-EC outperforms Crisp-EC for certain Gamma mean values. For high Gamma mean values, Prob-EC and Crisp-EC are equally inaccurate. In the case of `intermediate' noise, there are at least three probabilistic conjuncts per initiation condition. Due to the fact that fights have a relatively short duration (much shorter than \meet, for example), there are not enough initiations to raise Prob-EC's probability above the 0.5 threshold in high Gamma mean values.

Regarding \leave\ (see Figure \ref{fig:exp-original-intermediate-fmeasure-05}(d)), Crisp-EC seems to fair slightly better than Prob-EC for low Gamma mean values. Under the `intermediate' noise assumption, Prob-EC has to consider more probabilistic conjuncts per initiation condition, due to the presence of \close. As a result, and given the single initiation of \leave, Prob-EC tends to produce probabilities below 0.5 for this LTA, even in cases where the STA probabilities themselves might be above 0.5 and therefore sufficient to allow Crisp-EC to recognise \leave. For higher Gamma mean values, Prob-EC and Crisp-EC are equally inaccurate, because the data required by Crisp-EC to initiate \leave, whether it is the \person\ fluent, the \inactive\ STA or the coordinate fluents of the entities involved, are missing. 

\begin{figure}[H]
        \centering
        \begin{subfigure}[b]{.5\textwidth}
                \centering
                \includegraphics[width=\textwidth]{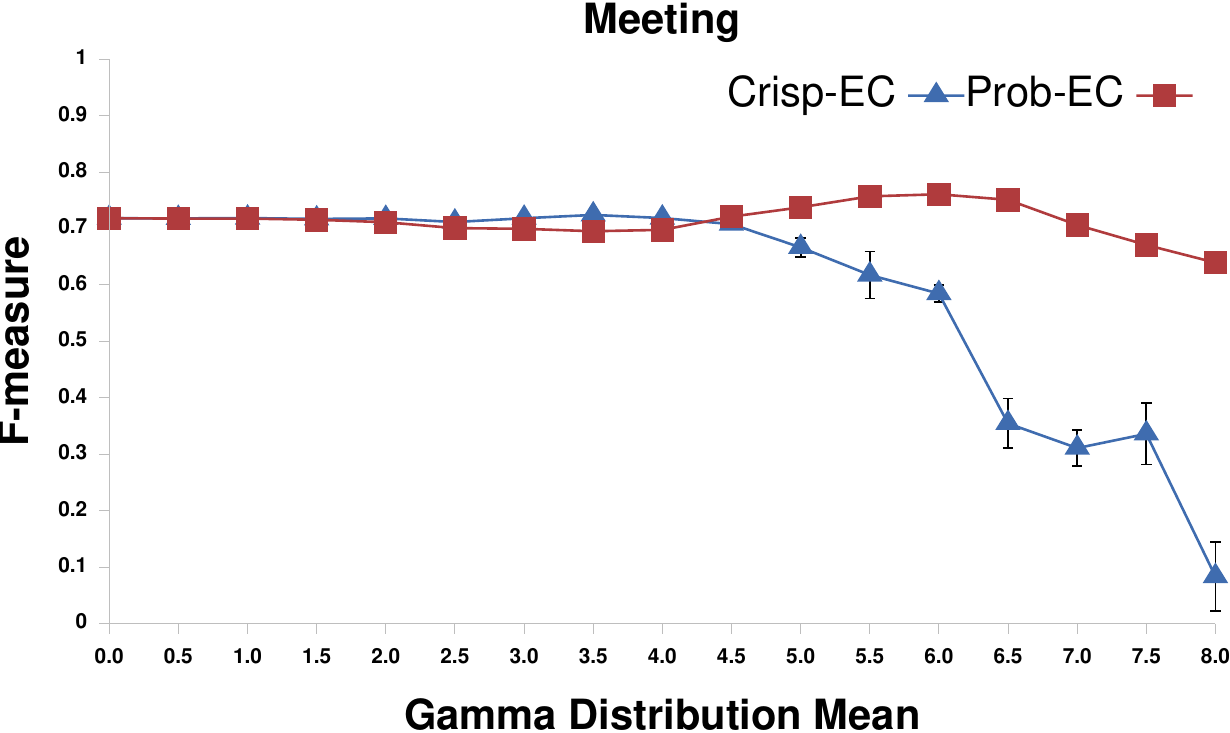}
                \caption{Threshold$\val$0.3}
        \end{subfigure}%
        ~ 
        \begin{subfigure}[b]{.5\textwidth}
                \centering
                \includegraphics[width=\textwidth]{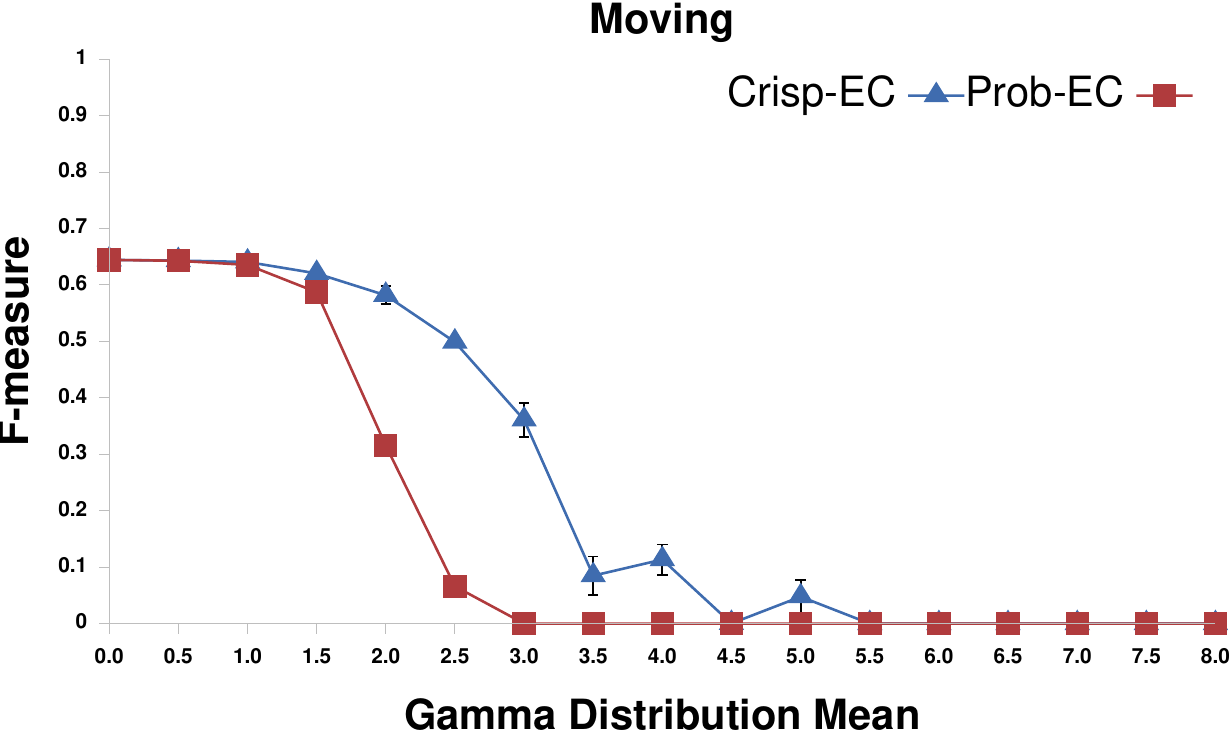}
                \caption{Threshold$\val$0.3}
        \end{subfigure}
        \linebreak\linebreak 
        \begin{subfigure}[b]{.5\textwidth}
                \centering
                \includegraphics[width=\textwidth]{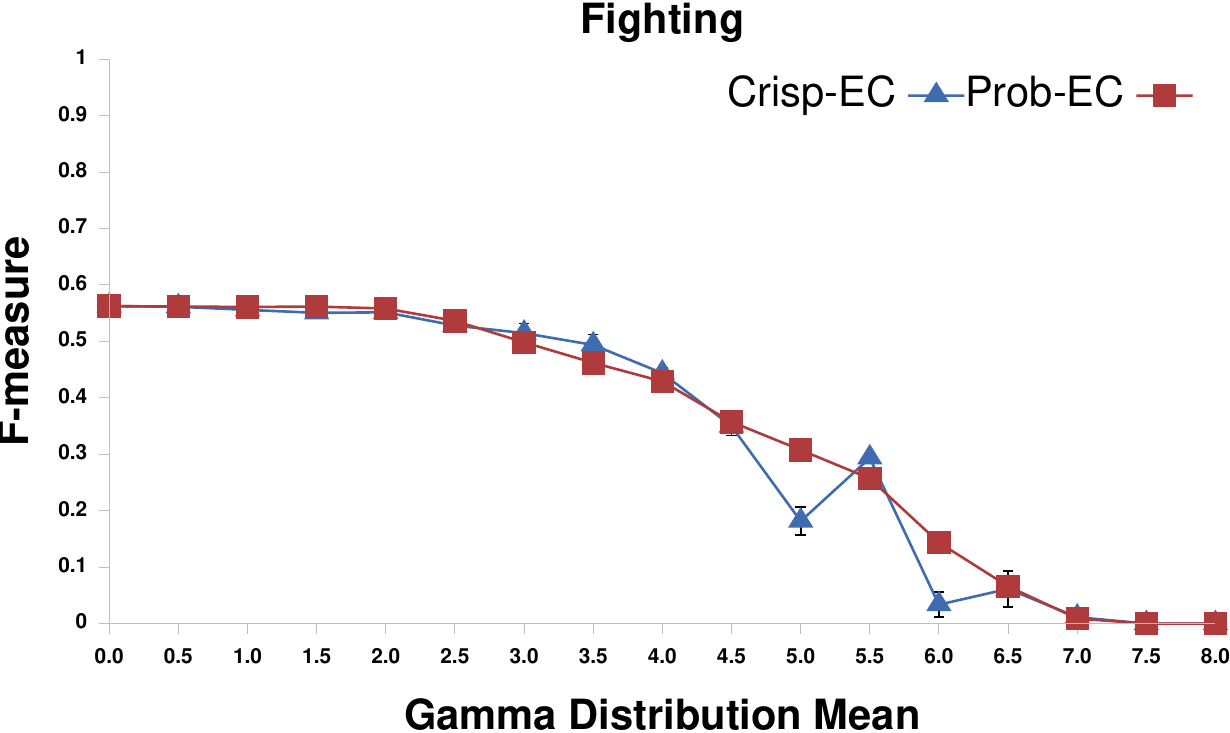}
                \caption{Threshold$\val$0.3}
        \end{subfigure}~
	\begin{subfigure}[b]{.5\textwidth}
                \centering
                \includegraphics[width=\textwidth]{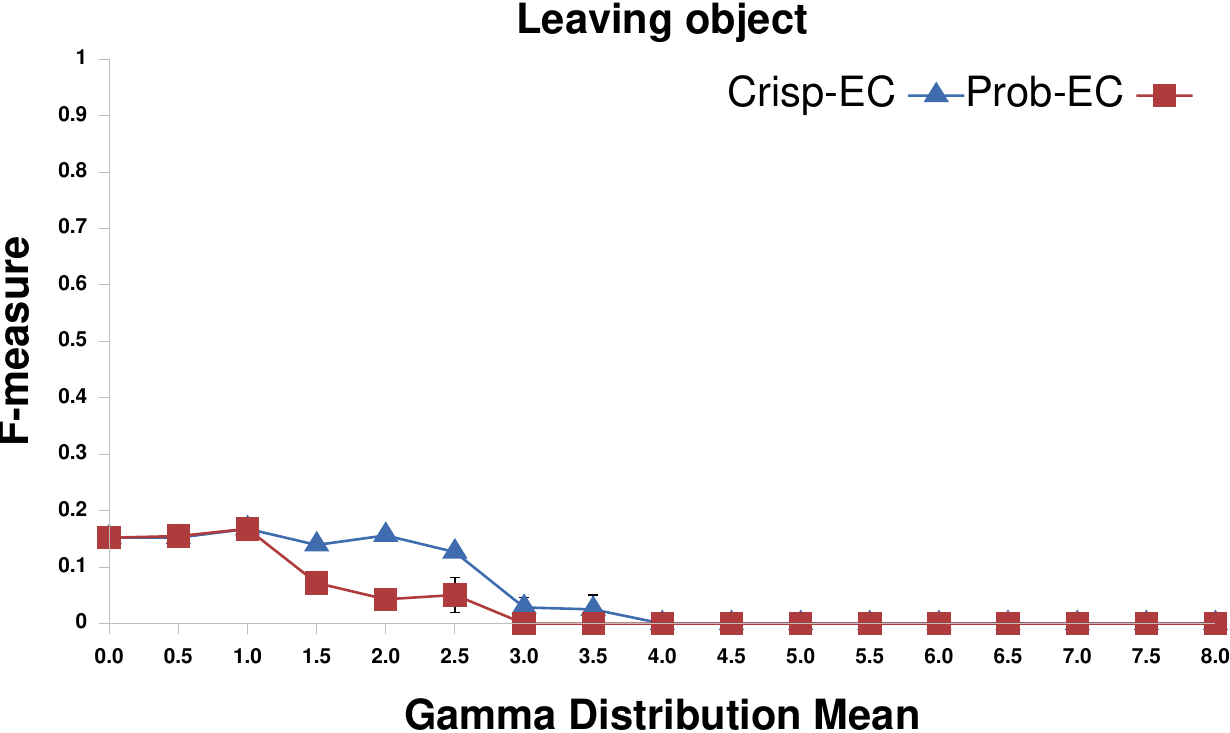}
                \caption{Threshold$\val$0.3}
        \end{subfigure}
	\linebreak\linebreak
        \begin{subfigure}[b]{.5\textwidth}
                \centering
                \includegraphics[width=\textwidth]{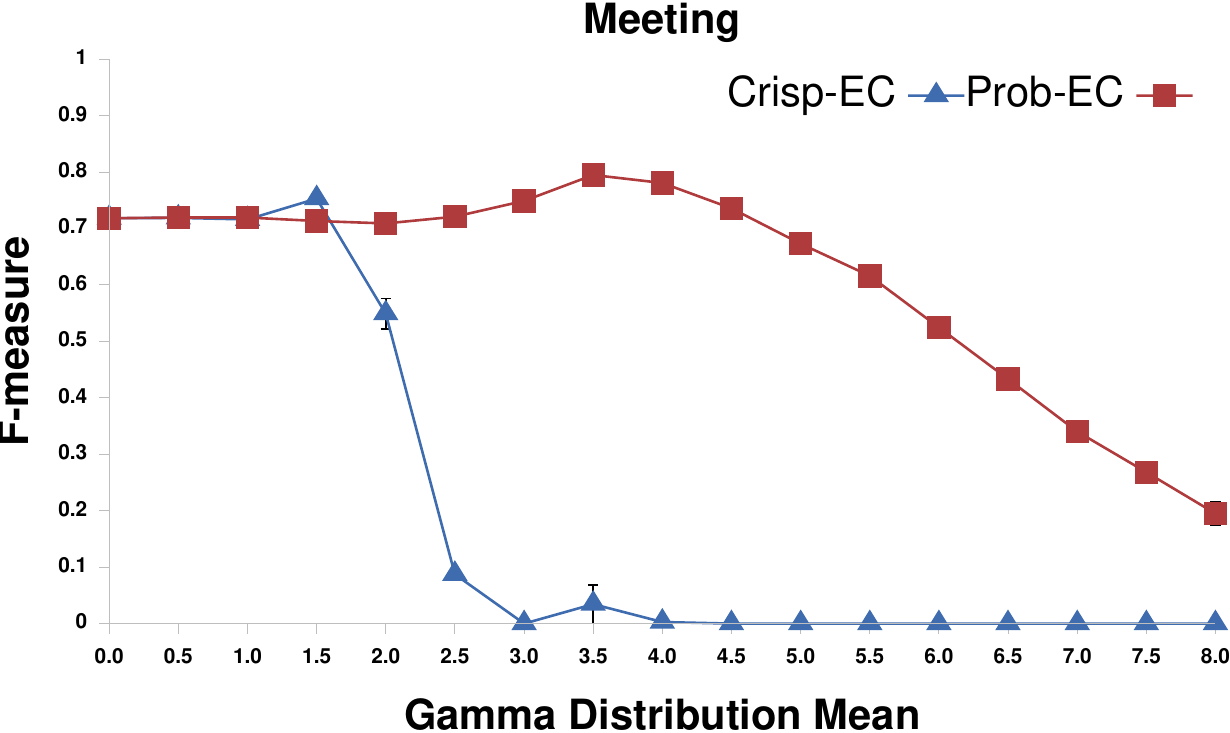}
                \caption{Threshold$\val$0.7}
        \end{subfigure}%
        ~ 
        \begin{subfigure}[b]{.5\textwidth}
                \centering
                \includegraphics[width=\textwidth]{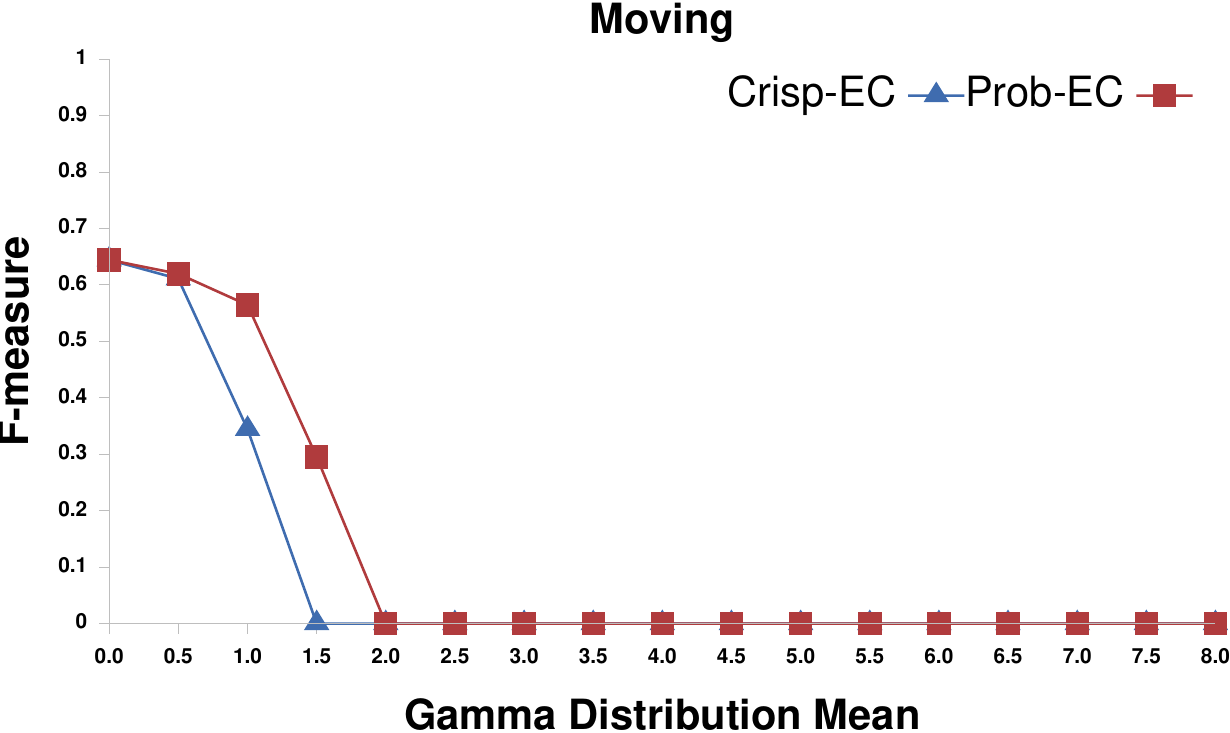}
                \caption{Threshold$\val$0.7}
        \end{subfigure}
        \linebreak\linebreak 
        \begin{subfigure}[b]{.5\textwidth}
                \centering
                \includegraphics[width=\textwidth]{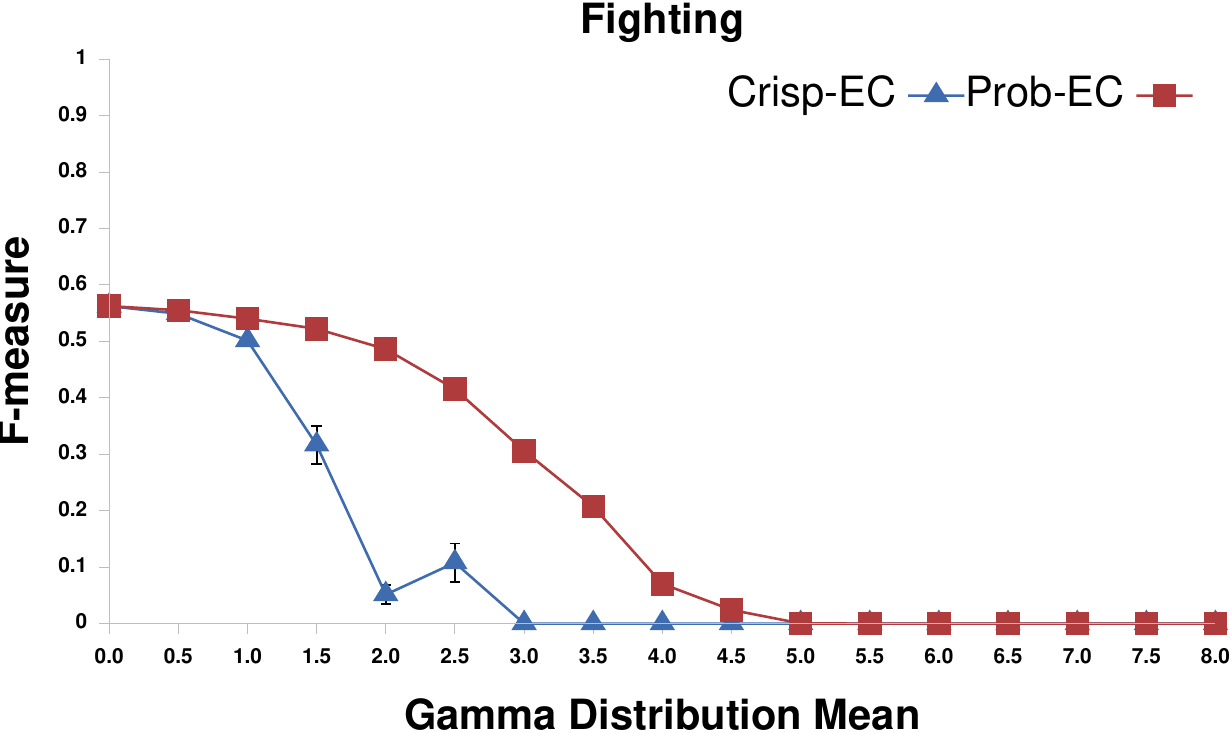}
                \caption{Threshold$\val$0.7}
        \end{subfigure}~
	\begin{subfigure}[b]{.5\textwidth}
                \centering
                \includegraphics[width=\textwidth]{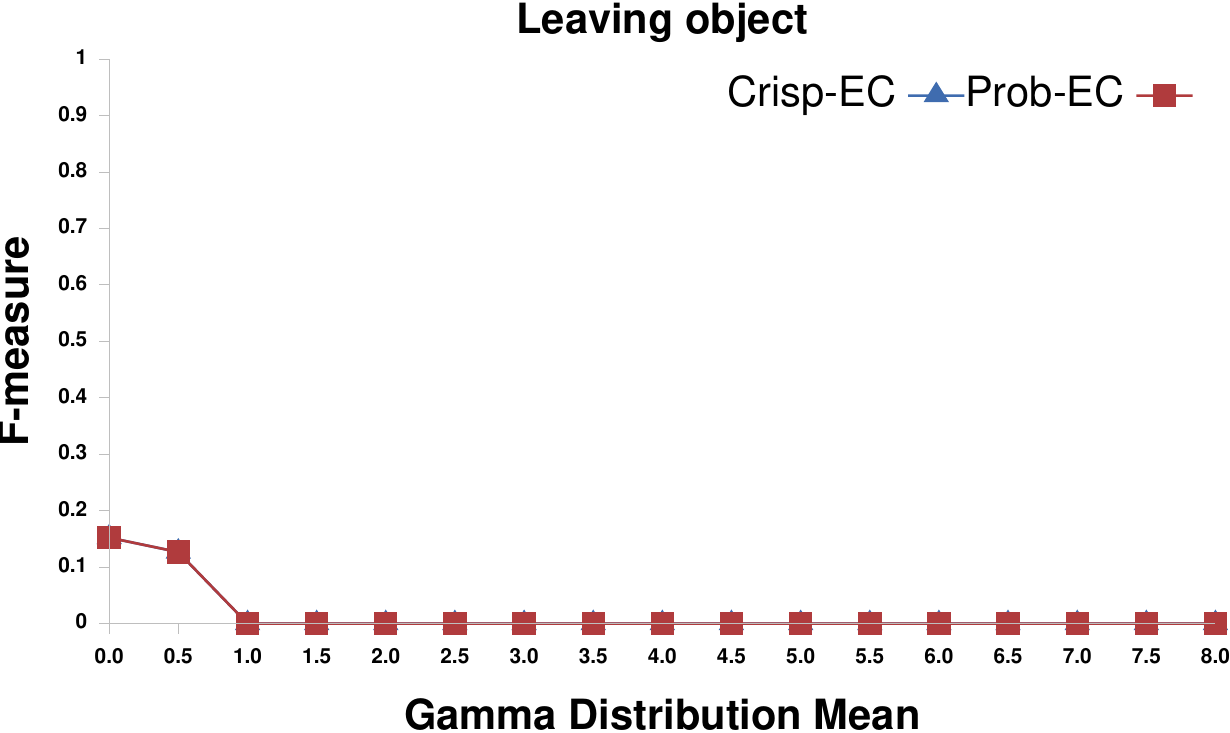}
                \caption{Threshold$\val$0.7}
        \end{subfigure}
        \caption{Crisp-EC and Prob-EC F-measure per Gamma mean value under `intermediate' noise. In the first four figures the threshold is 0.3 while in the last four figures the threshold is 0.7.}\label{fig:exp-original-intermediate-fmeasure-03-07}
\end{figure}

Figure \ref{fig:exp-original-intermediate-fmeasure-03-07} compares the recognition accuracy of Crisp-EC and Prob-EC using 0.3 and 0.7 thresholds. As in the case of the `smooth' noise experiments, Crisp-EC is affected more than Prob-EC by the threshold change --- the accuracy of Crisp-EC improves as the threshold decreases (the lower the threshold, the closer the input of Crisp-EC to the original, artificial noise-free dataset).

\begin{figure}[h]
        \centering
        \begin{subfigure}[b]{.5\textwidth}
                \centering
                \includegraphics[width=\textwidth]{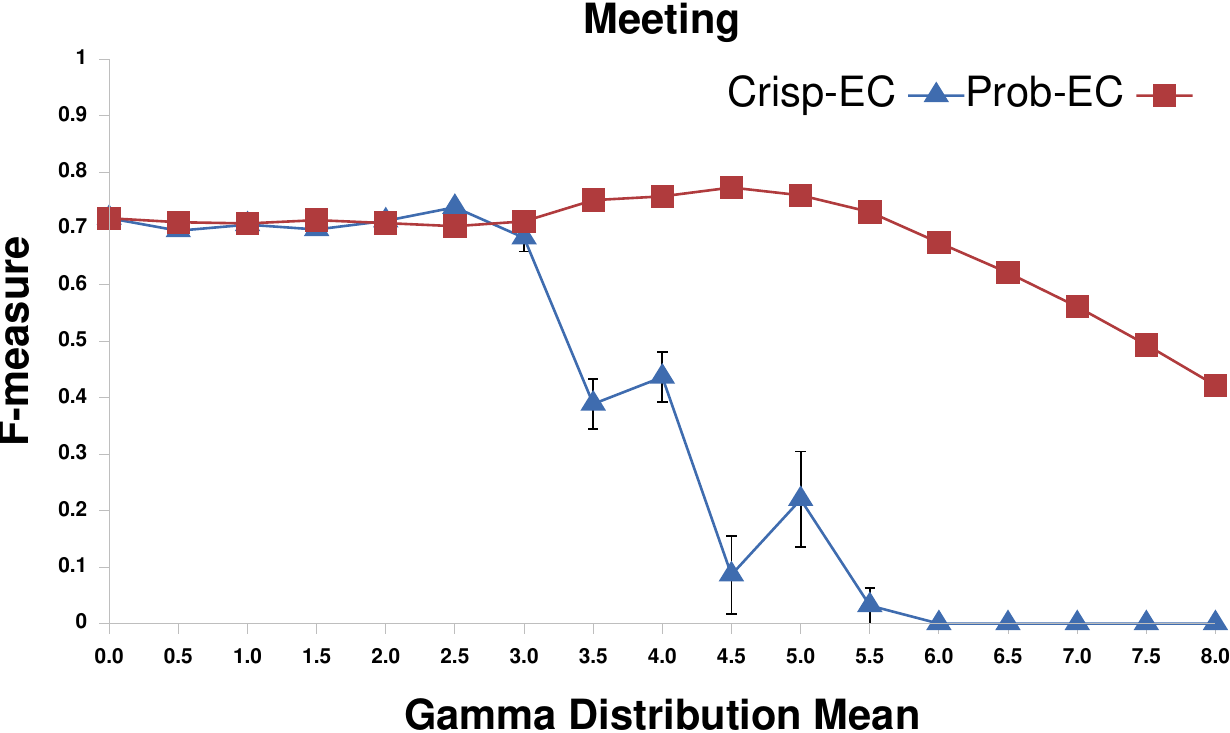}
                \caption{(a)}
                \label{fig:exp-original-strong-fmeasure-05-meeting}
        \end{subfigure}%
        ~ 
        \begin{subfigure}[b]{.5\textwidth}
                \centering
                \includegraphics[width=\textwidth]{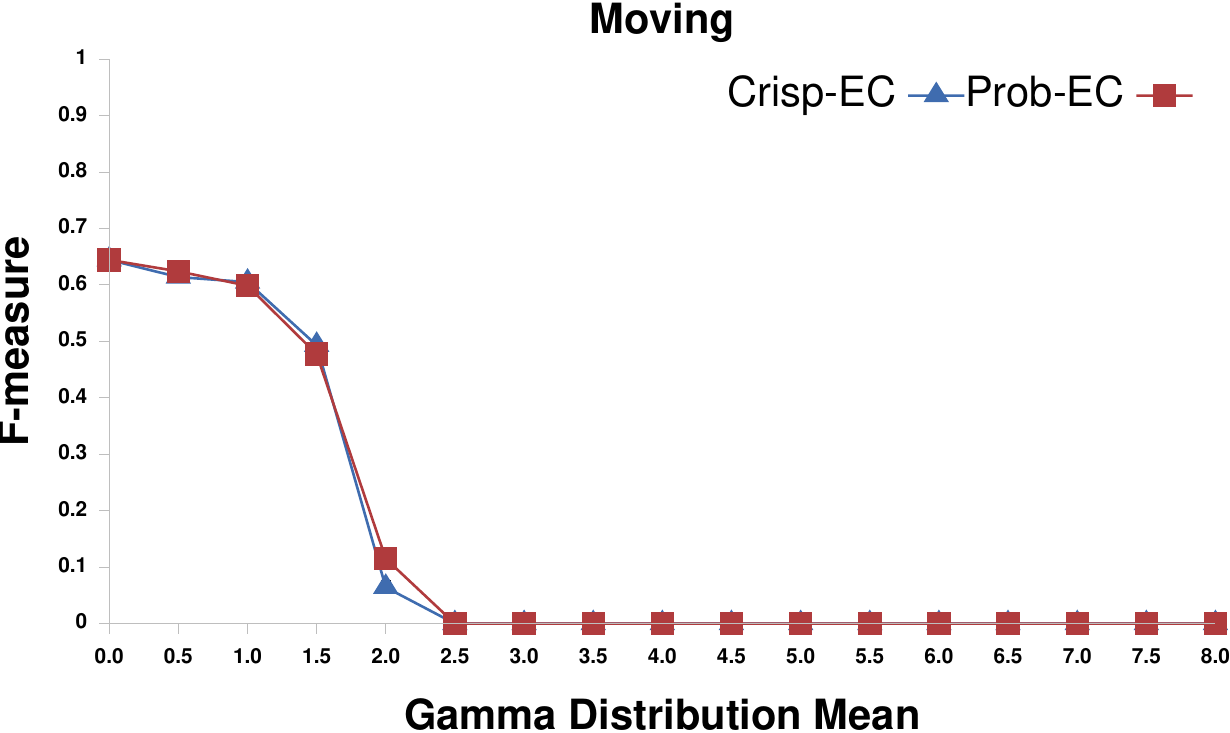}
                \caption{(b)}
                \label{fig:exp-original-strong-fmeasure-05-moving}
        \end{subfigure}
        \linebreak\linebreak 
        \begin{subfigure}[b]{.5\textwidth}
                \centering
                \includegraphics[width=\textwidth]{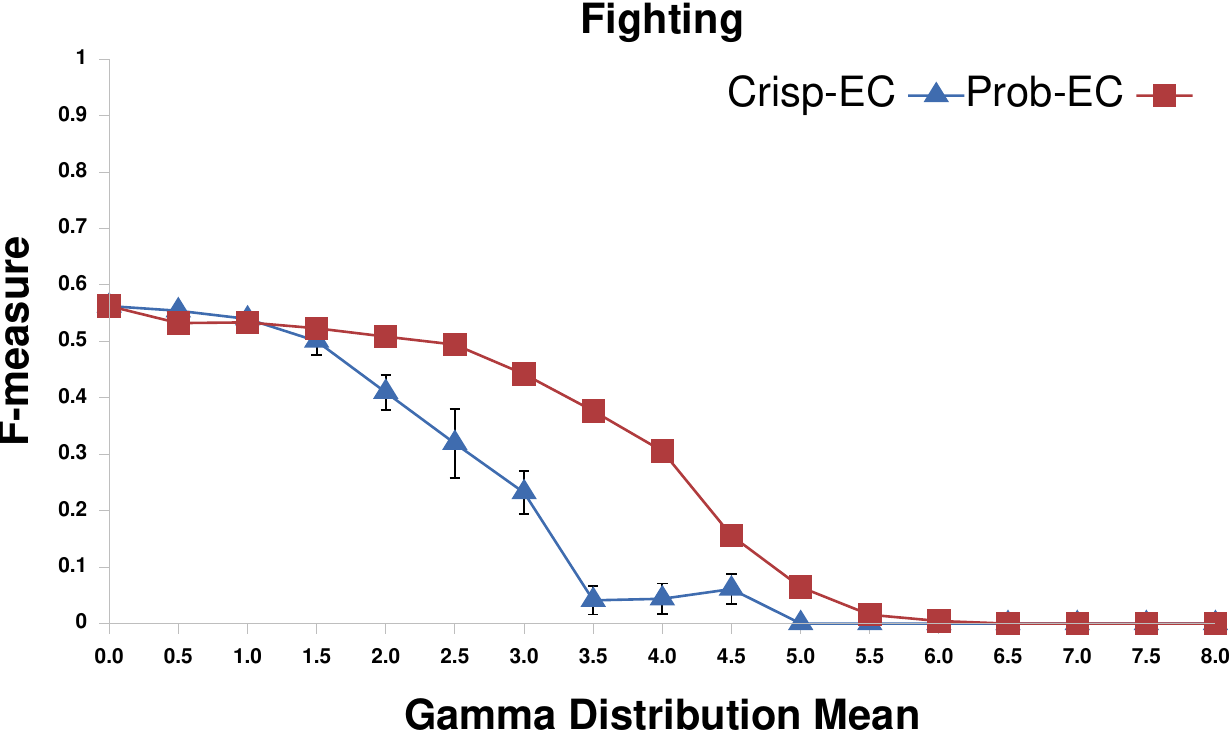}
                \caption{(c)}
                \label{fig:exp-original-strong-fmeasure-05-fighting}
        \end{subfigure}~
	\begin{subfigure}[b]{.5\textwidth}
                \centering
                \includegraphics[width=\textwidth]{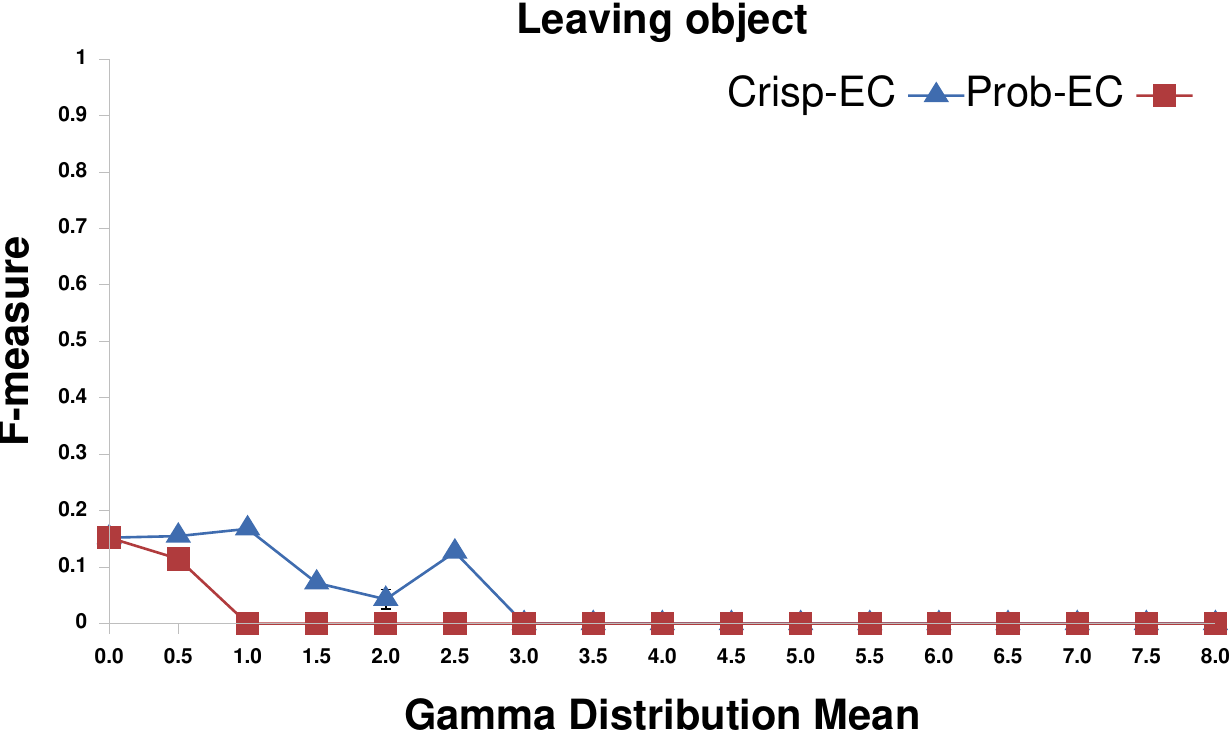}
                \caption{(d)}
                \label{fig:exp-original-strong-fmeasure-05-leaving-object}
        \end{subfigure}
        \caption{Crisp-EC and Prob-EC F-measure per Gamma mean value under `strong' noise and a 0.5 threshold.}\label{fig:exp-original-strong-fmeasure-05}
\end{figure}

\subsubsection{`Strong' Noise Experiments}\label{subsec:strongNoiseExps}


Figure \ref{fig:exp-original-strong-fmeasure-05} compares the recognition accuracy of Crisp-EC and Prob-EC  under `strong' noise using a 0.5 threshold. As can be seen from this figure, the recognition accuracy under `strong' noise is very similar to that of `intermediate' noise in the 0.5 threshold. For Prob-EC, we need a series of spurious information to surpass the chosen threshold --- the introduction of spurious STA in the `strong' noise experiments is not that systematic and does not create problems to Prob-EC. Crisp-EC is very slightly affected by the introduction of spurious STA: it now has a number of additional FP, mainly in the case of \move.
Even a single initiation condition fired by a spurious \walking\ STA and some other CAVIAR STA may be enough in Crisp-EC to create a series of FP. However, in low Gamma values most of the spurious STA have low probabilities --- below the 0.5 threshold --- and therefore are not part of the input of Crisp-EC. Furthermore, in high Gamma values the spurious STA cannot initiate a LTA because several of the CAVIAR STA tend to be deleted from the input of Crisp-EC as they have low probabilities. 

\begin{figure}[t]
        \centering
        \begin{subfigure}[b]{.5\textwidth}
                \centering
                \includegraphics[width=\textwidth]{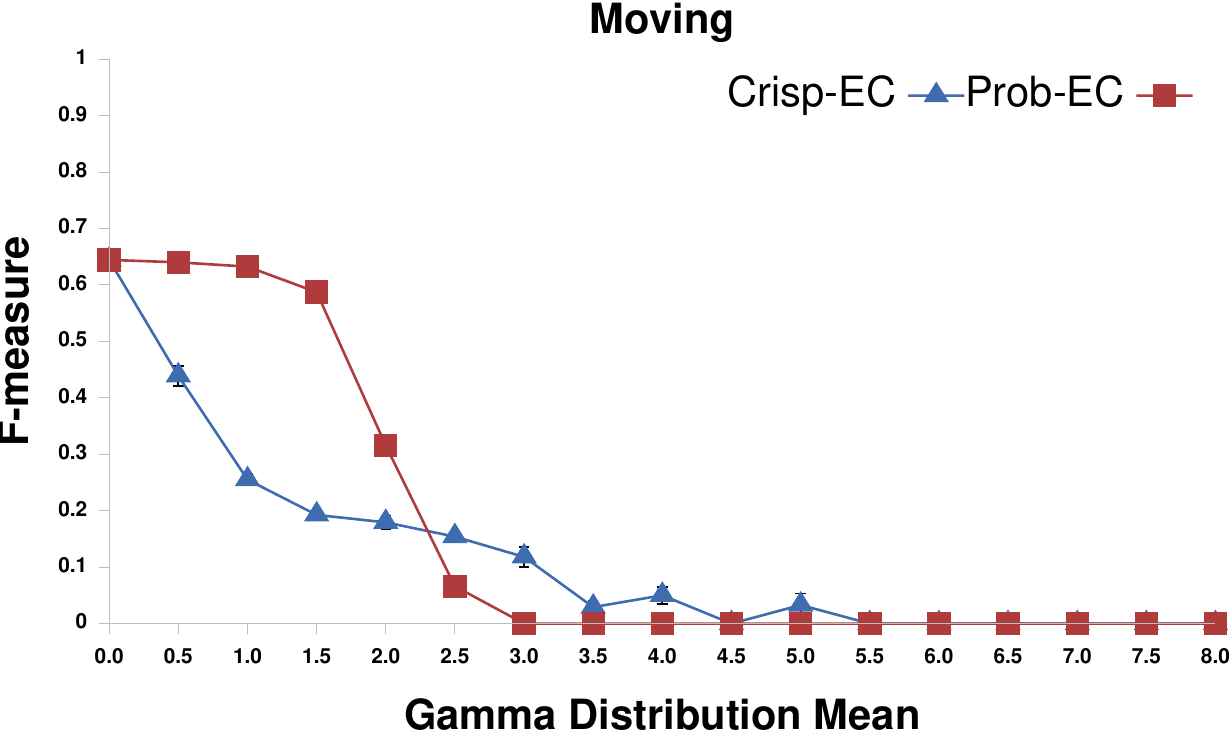}
	      \caption{Threshold$\val$0.3}
        \end{subfigure}~
        \begin{subfigure}[b]{.5\textwidth}
                \centering
                \includegraphics[width=\textwidth]{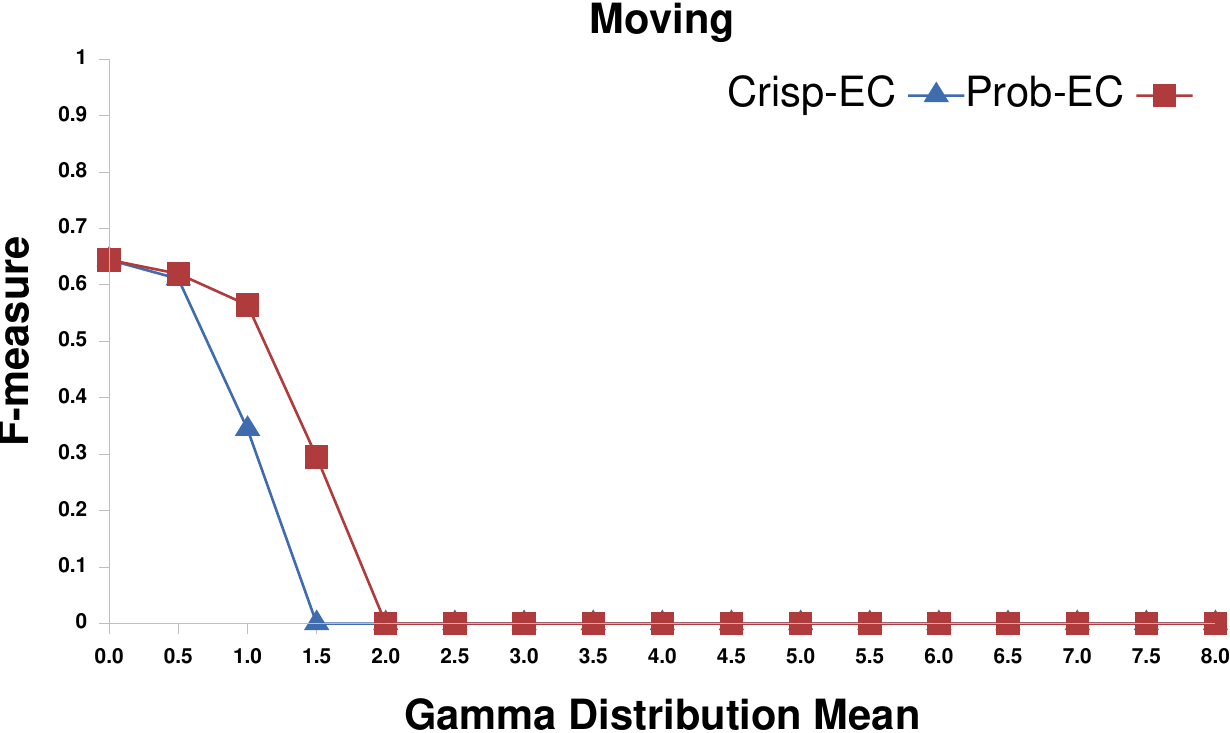}
                \caption{Threshold$\val$0.7}
        \end{subfigure}
        \caption{Crisp-EC and Prob-EC F-measure per Gamma mean value under `strong' noise in \move.}\label{fig:exp-original-strong-fmeasure-03-07}
\end{figure}

Figure \ref{fig:exp-original-strong-fmeasure-03-07} compares the accuracy of Crisp-EC and Prob-EC in \move\ using thresholds of 0.3 and 0.7 (the diagrams for the remaining LTA are omitted as they are similar to those concerning `intermediate' noise). Crisp-EC, as in the previous experimental settings, is more significantly affected than Prob-EC by the threshold change. Unlike the last two experimental settings, reducing the threshold causes problems to Crisp-EC with respect to \move. With a lower threshold, Crisp-EC loses less CAVIAR STA, but at the same time keeps more spurious facts, creating many more FP. Prob-EC, even in the low threshold of 0.3, is almost unaffected by the spurious facts.
In the 0.7 threshold Crisp-EC has a very small number of spurious facts in its input and, therefore, its accuracy is not compromised by these facts.

In addition to \walking\ STA, we could have added other spurious facts such \abrupt, \activeb, or \inactive\ and repeated the experiments. The results would have been similar to those presented above --- for example, spurious \abrupt\ STA would have compromised the accuracy of Crisp-EC in \fight, especially in low thresholds.

\section{Summary and Future Work}\label{sec:summ} 

We presented Prob-EC, an Event Calculus dialect for probabilistic reasoning. 
Prob-EC is the first Event Calculus dialect able to deal with uncertainty in the input STA. Moreover, this is the first approach that thoroughly evaluates the Event Calculus in a probabilistic framework. 
Our experimental evaluation on a benchmark activity recognition dataset showed that Prob-EC outperforms Crisp-EC when:
\begin{itemize}
 \item a LTA has multiple initiations, and
 \item the LTA depends on a small number of probabilistic conjuncts.
\end{itemize}
%

When a LTA depends on a large number of probabilistic conjuncts, Prob-EC is significantly affected and its performance is close to that of Crisp-EC. ProbLog makes an independence assumption about input facts and thus the product of the probabilities of many probabilistic conjuncts may be very small, even if the probability of each individual conjunct is high. The greater the number of probabilistic conjuncts, the more initiations Prob-EC requires to surpass the chosen recognition threshold. 

Prob-EC is more resilient to spurious facts than Crisp-EC. Even a single such fact may create a series of false positives (FP) in Crisp-EC, whereas this type of noise must be much more systematic to affect Prob-EC: to impinge the accuracy of the latter we need a series of spurious facts in order to surpass the recognition threshold.

In the case of a single initiation, there are situations in which Prob-EC may fare significantly better than Crisp-EC and vice versa, but these did not arise in our experiments. In general, Prob-EC is expected to have similar performance to Crisp-EC on LTA with a single initiation in the case of a small number of probabilistic conjuncts, while Crisp-EC is likely to perform better in the case of a large number of such conjuncts.

The experimental results concerning one or more terminations are similar to those regarding one or more initiations. 
For example, Prob-EC outperforms Crisp-EC in the case of LTA with multiple terminations that depend on a small number of probabilistic conjuncts. The repeated terminations allow Prob-EC to stop recognising a LTA when Crisp-EC continues the recognition producing FP.


There are several directions for further work. 
First, we intend to experiment with additional types of noise and additional datasets.
Second, we aim to incorporate advanced caching techniques in Prob-EC, such as those presented in (Artikis, Sergot and Paliouras \citeyear{artikisDEBS12}), in order to make it suitable for run-time activity recognition. 
Finally, we aim to address the issue of imprecise LTA definitions --- we intend to provide a unified framework that will be able to deal with both STA detection probabilities and imperfect LTA definitions.

\section*{Acknowledgements}

This work was supported by the EU PRONTO Project (FP7-ICT 231738). We are very grateful to the ProbLog developing team whose feedback contributed greatly to our grasp of the language. The authors themselves, however, are solely responsible for any misunderstanding about its use. We have also benefited from discussions with Marek Sergot on the Event Calculus.  

\bibliographystyle{acmtrans}

\begin{thebibliography}{}

\bibitem[\protect\citeauthoryear{Allen}{Allen}{1983}]{Allen83}
{\sc Allen, J.} 1983.
\newblock Maintaining knowledge about temporal intervals.
\newblock {\em Communications of the {ACM}\/}~{\em 26,\/}~11, 832--843.

\bibitem[\protect\citeauthoryear{Artikis, Sergot, and Paliouras}{Artikis
  et~al\mbox{.}}{2010}]{artikis10EIMM}
{\sc Artikis, A.}, {\sc Sergot, M.}, {\sc and} {\sc Paliouras, G.} 2010.
\newblock A logic programming approach to activity recognition.
\newblock In {\em Proceedings of ACM Workshop on Events in Multimedia}.

\bibitem[\protect\citeauthoryear{Artikis, Sergot, and Paliouras}{Artikis
  et~al\mbox{.}}{2012}]{artikisDEBS12}
{\sc Artikis, A.}, {\sc Sergot, M.}, {\sc and} {\sc Paliouras, G.} 2012.
\newblock Run-time composite event recognition.
\newblock In {\em Proceedings of International Conference on Distributed
  Event-Based Systems (DEBS)}. ACM, 69--80.

\bibitem[\protect\citeauthoryear{Artikis, Skarlatidis, Portet, and
  Paliouras}{Artikis et~al\mbox{.}}{2012}]{artikisKER}
{\sc Artikis, A.}, {\sc Skarlatidis, A.}, {\sc Portet, F.}, {\sc and} {\sc
  Paliouras, G.} 2012.
\newblock Logic-based event recognition.
\newblock {\em Knowledge Engineering Review\/}~{\em 27,\/}~4, 469--506.

\bibitem[\protect\citeauthoryear{Biswas, Thrun, and Fujimura}{Biswas
  et~al\mbox{.}}{2007}]{BiswasTF07}
{\sc Biswas, R.}, {\sc Thrun, S.}, {\sc and} {\sc Fujimura, K.} 2007.
\newblock Recognizing activities with multiple cues.
\newblock In {\em Proceedings of Workshop on Human Motion}, {A.~M. Elgammal},
  {B.~Rosenhahn}, {and} {R.~Klette}, Eds. LNCS, vol. 4814. Springer, 255--270.

\bibitem[\protect\citeauthoryear{Brand, Oliver, and Pentland}{Brand
  et~al\mbox{.}}{1997}]{BrandOP97}
{\sc Brand, M.}, {\sc Oliver, N.}, {\sc and} {\sc Pentland, A.} 1997.
\newblock Coupled {H}idden {M}arkov {M}odels for complex action recognition.
\newblock In {\em Proceedings of International Conference on Computer Vision
  and Pattern Recognition (CVPR)}. IEEE Computer Society, 994--999.

\bibitem[\protect\citeauthoryear{Brendel, Fern, and Todorovic}{Brendel
  et~al\mbox{.}}{2011}]{brendel2011PEL}
{\sc Brendel, W.}, {\sc Fern, A.}, {\sc and} {\sc Todorovic, S.} 2011.
\newblock Probabilistic event logic for interval-based event recognition.
\newblock In {\em Proceedings of International Conference on Computer Vision
  and Pattern Recognition (CVPR)}. IEEE, 3329--3336.

\bibitem[\protect\citeauthoryear{Bruynooghe, Mantadelis, Kimmig, Gutmann,
  Vennekens, Janssens, and Raedt}{Bruynooghe et~al\mbox{.}}{2010}]{foproblog}
{\sc Bruynooghe, M.}, {\sc Mantadelis, T.}, {\sc Kimmig, A.}, {\sc Gutmann,
  B.}, {\sc Vennekens, J.}, {\sc Janssens, G.}, {\sc and} {\sc Raedt, L.~D.}
  2010.
\newblock Prob{L}og technology for inference in a probabilistic first order
  logic.
\newblock In {\em Proceedings of European Conference on Artificial Intelligence
  (ECAI)}. 719--724.

\bibitem[\protect\citeauthoryear{Bryant}{Bryant}{1986}]{bdd}
{\sc Bryant, R.} 1986.
\newblock Graph-based algorithms for boolean function manipulation.
\newblock {\em IEEE Transactions on Computers\/}~{\em 35(8)}, 677--691.

\bibitem[\protect\citeauthoryear{Cugola and Margara}{Cugola and
  Margara}{2012}]{cugola11}
{\sc Cugola, G.} {\sc and} {\sc Margara, A.} 2012.
\newblock Processing flows of information: From data stream to complex event
  processing.
\newblock {\em ACM Computing Surveys\/}~{\em 44,\/}~3, 15.

\bibitem[\protect\citeauthoryear{Dousson and Maigat}{Dousson and
  Maigat}{2007}]{dousson07}
{\sc Dousson, C.} {\sc and} {\sc Maigat, P.~L.} 2007.
\newblock Chronicle recognition improvement using temporal focusing and
  hierarchisation.
\newblock In {\em Proceedings of International Joint Conference on Artificial
  Intelligence (IJCAI)}. 324--329.

\bibitem[\protect\citeauthoryear{Fierens, den Broeck, Thon, Gutmann, and
  Raedt}{Fierens et~al\mbox{.}}{2011}]{weightedcnfs}
{\sc Fierens, D.}, {\sc den Broeck, G.~V.}, {\sc Thon, I.}, {\sc Gutmann, B.},
  {\sc and} {\sc Raedt, L.~D.} 2011.
\newblock Inference in probabilistic logic programs using weighted {CNF}'s.
\newblock In {\em Proceedings of International Conference on Uncertainty in
  Artificial Intelligence (UAI)}. 211--220.

\bibitem[\protect\citeauthoryear{Gibson}{Gibson}{1979}]{gibson1979ecological}
{\sc Gibson, J.} 1979.
\newblock The ecological approach to visual perception.

\bibitem[\protect\citeauthoryear{Ginsberg}{Ginsberg}{1990}]{ginsberg90}
{\sc Ginsberg, M.~L.} 1990.
\newblock Bilattices and modal operators.
\newblock {\em Journal of Logic and Computation\/}~{\em 1}, 1--41.

\bibitem[\protect\citeauthoryear{Gong and Xiang}{Gong and
  Xiang}{2003}]{gong2003}
{\sc Gong, S.} {\sc and} {\sc Xiang, T.} 2003.
\newblock Recognition of group activities using dynamic probabilistic networks.
\newblock In {\em Proceedings of International Conference on Computer Vision}.
  IEEE, 742--749.

\bibitem[\protect\citeauthoryear{Hakeem and Shah}{Hakeem and
  Shah}{2007}]{shah07AIJ}
{\sc Hakeem, A.} {\sc and} {\sc Shah, M.} 2007.
\newblock Learning, detection and representation of multi-agent events in
  videos.
\newblock {\em Artificial Intelligence\/}~{\em 171,\/}~8--9, 586--605.

\bibitem[\protect\citeauthoryear{Helaoui, Niepert, and Stuckenschmidt}{Helaoui
  et~al\mbox{.}}{2011}]{HelaouiNS11}
{\sc Helaoui, R.}, {\sc Niepert, M.}, {\sc and} {\sc Stuckenschmidt, H.} 2011.
\newblock Recognizing interleaved and concurrent activities: A
  statistical-relational approach.
\newblock In {\em Proceedings of International Conference on Pervasive
  Computing and Communications}. IEEE, 1--9.

\bibitem[\protect\citeauthoryear{Hongeng and Nevatia}{Hongeng and
  Nevatia}{2003}]{hongeng2003large}
{\sc Hongeng, S.} {\sc and} {\sc Nevatia, R.} 2003.
\newblock Large-scale event detection using semi-{H}idden {M}arkov {M}odels.
\newblock In {\em Proceedings of International Conference on Computer Vision}.
  IEEE, 1455--1462.

\bibitem[\protect\citeauthoryear{Kembhavi, Yeh, and Davis}{Kembhavi
  et~al\mbox{.}}{2010}]{davis2010}
{\sc Kembhavi, A.}, {\sc Yeh, T.}, {\sc and} {\sc Davis, L.~S.} 2010.
\newblock Why did the person cross the road (there)? scene understanding using
  probabilistic logic models and common sense reasoning.
\newblock In {\em Proceedings of European Conference on Computer Vision
  (ECCV)}. 693--706.

\bibitem[\protect\citeauthoryear{Kimmig, Demoen, Raedt, Costa, and
  Rocha}{Kimmig et~al\mbox{.}}{2011}]{kimmig11}
{\sc Kimmig, A.}, {\sc Demoen, B.}, {\sc Raedt, L.~D.}, {\sc Costa, V.~S.},
  {\sc and} {\sc Rocha, R.} 2011.
\newblock On the implementation of the probabilistic logic programming language
  {P}rob{L}og.
\newblock {\em Theory and Practice of Logic Programming\/}~{\em 11}, 235--262.

\bibitem[\protect\citeauthoryear{Kosmopoulos, Antonakaki, Valasoulis, Kesidis,
  and Perantonis}{Kosmopoulos et~al\mbox{.}}{2008}]{kosmo08setn}
{\sc Kosmopoulos, D.}, {\sc Antonakaki, P.}, {\sc Valasoulis, K.}, {\sc
  Kesidis, A.}, {\sc and} {\sc Perantonis, S.} 2008.
\newblock Human behaviour classification using multiple views.
\newblock In {\em Proceedings of Hellenic Conference on Artificial
  Intelligence}. Vol. 5138. Springer.

\bibitem[\protect\citeauthoryear{Kowalski and Sergot}{Kowalski and
  Sergot}{1986}]{kowalski86}
{\sc Kowalski, R.} {\sc and} {\sc Sergot, M.} 1986.
\newblock A logic-based calculus of events.
\newblock {\em New Generation Computing\/}~{\em 4,\/}~1, 67--96.

\bibitem[\protect\citeauthoryear{Lafferty, McCallum, and Pereira}{Lafferty
  et~al\mbox{.}}{2001}]{LaffertyMP01CRF}
{\sc Lafferty, J.~D.}, {\sc McCallum, A.}, {\sc and} {\sc Pereira, F. C.~N.}
  2001.
\newblock Conditional random fields: Probabilistic models for segmenting and
  labeling sequence data.
\newblock In {\em Proceedings of International Conference on Machine Learning
  (ICML)}, {C.~E. Brodley} {and} {A.~P. Danyluk}, Eds. Morgan Kaufmann,
  282--289.

\bibitem[\protect\citeauthoryear{Liao, Fox, and Kautz}{Liao
  et~al\mbox{.}}{2007}]{liao2007hierarchical}
{\sc Liao, L.}, {\sc Fox, D.}, {\sc and} {\sc Kautz, H.} 2007.
\newblock Hierarchical conditional random fields for {GPS}-based activity
  recognition.
\newblock {\em Robotics Research\/}, 487--506.

\bibitem[\protect\citeauthoryear{Luckham}{Luckham}{2002}]{luckhamBook}
{\sc Luckham, D.} 2002.
\newblock {\em The Power of Events: An Introduction to Complex Event Processing
  in Distributed Enterprise Systems}.
\newblock Addison-Wesley.

\bibitem[\protect\citeauthoryear{Moldovan, Moreno, van Otterlo, Santos-Victor,
  and De~Raedt}{Moldovan et~al\mbox{.}}{2012}]{Moldovan12a}
{\sc Moldovan, B.}, {\sc Moreno, P.}, {\sc van Otterlo, M.}, {\sc
  Santos-Victor, J.}, {\sc and} {\sc De~Raedt, L.} 2012.
\newblock Learning relational affordance models for robots in multi-object
  manipulation tasks.
\newblock In {\em Proceedings of International Conference on Robotics and
  Automation (ICRA)}. 4373--4378.

\bibitem[\protect\citeauthoryear{Morariu and Davis}{Morariu and
  Davis}{2011}]{morariu11}
{\sc Morariu, V.~I.} {\sc and} {\sc Davis, L.~S.} 2011.
\newblock Multi-agent event recognition in structured scenarios.
\newblock In {\em Proceedings of International Conference on Computer Vision
  and Pattern Recognition (CVPR)}. 3289--3296.

\bibitem[\protect\citeauthoryear{Murphy}{Murphy}{2002}]{murphy2002DBN}
{\sc Murphy, K.} 2002.
\newblock Dynamic bayesian networks: representation, inference and learning.
\newblock Ph.D. thesis, University of California.

\bibitem[\protect\citeauthoryear{Natarajan and Nevatia}{Natarajan and
  Nevatia}{2007}]{natarajan2007hierarchical}
{\sc Natarajan, P.} {\sc and} {\sc Nevatia, R.} 2007.
\newblock Hierarchical multi-channel semi {H}idden {M}arkov {M}odels.
\newblock In {\em Proceedings of International Joint Conference on Artificial
  Intelligence (IJCAI)}. 2562--2567.

\bibitem[\protect\citeauthoryear{Rabiner and Juang}{Rabiner and
  Juang}{1986}]{rabiner1986HMM}
{\sc Rabiner, L.} {\sc and} {\sc Juang, B.} 1986.
\newblock An introduction to {H}idden {M}arkov {M}odels.
\newblock {\em ASSP Magazine\/}~{\em 3,\/}~1, 4--16.

\bibitem[\protect\citeauthoryear{Richardson and Domingos}{Richardson and
  Domingos}{2006}]{mln2006}
{\sc Richardson, M.} {\sc and} {\sc Domingos, P.} 2006.
\newblock Markov logic networks.
\newblock {\em Machine Learning\/}~{\em 62,\/}~1-2, 107--136.

\bibitem[\protect\citeauthoryear{Sadilek and Kautz}{Sadilek and
  Kautz}{2012}]{sadilek2012}
{\sc Sadilek, A.} {\sc and} {\sc Kautz, H.} 2012.
\newblock Location-based reasoning about complex multi-agent behavior.
\newblock {\em Journal of Artificial Intelligence Research\/}~{\em 43},
  87--133.

\bibitem[\protect\citeauthoryear{Selman, Amer, Fern, and Todorovic}{Selman
  et~al\mbox{.}}{2011}]{selman2011PEL}
{\sc Selman, J.}, {\sc Amer, M.}, {\sc Fern, A.}, {\sc and} {\sc Todorovic, S.}
  2011.
\newblock {PEL}-{CNF}: Probabilistic event logic conjunctive normal form for
  video interpretation.
\newblock In {\em Computer Vision Workshops}. IEEE, 680--687.

\bibitem[\protect\citeauthoryear{Shet, Harwood, and Davis}{Shet
  et~al\mbox{.}}{2005}]{davis05AVSS}
{\sc Shet, V.}, {\sc Harwood, D.}, {\sc and} {\sc Davis, L.} 2005.
\newblock {V}id{MAP}: {v}ideo monitoring of activity with {P}rolog.
\newblock In {\em Proceedings of International Conference on Advanced Video and
  Signal Based Surveillance (AVSS)}. IEEE, 224--229.

\bibitem[\protect\citeauthoryear{Shet, Neumann, Ramesh, and Davis}{Shet
  et~al\mbox{.}}{2007}]{davis07CVPR}
{\sc Shet, V.}, {\sc Neumann, J.}, {\sc Ramesh, V.}, {\sc and} {\sc Davis, L.}
  2007.
\newblock Bilattice-based logical reasoning for human detection.
\newblock In {\em Proceedings of International Conference on Computer Vision
  and Pattern Recognition (CVPR)}. IEEE, 1--8.

\bibitem[\protect\citeauthoryear{Siskind}{Siskind}{2001}]{siskind2001EL}
{\sc Siskind, J.} 2001.
\newblock Grounding the lexical semantics of verbs in visual perception using
  force dynamics and event logic.
\newblock {\em Journal of Artificial Intelligence Research\/}~{\em 15}, 31--90.

\bibitem[\protect\citeauthoryear{Skarlatidis, Paliouras, Vouros, and
  Artikis}{Skarlatidis et~al\mbox{.}}{2011}]{anskarlRuleML}
{\sc Skarlatidis, A.}, {\sc Paliouras, G.}, {\sc Vouros, G.~A.}, {\sc and} {\sc
  Artikis, A.} 2011.
\newblock Probabilistic event calculus based on markov logic networks.
\newblock In {\em RuleML America}. 155--170.

\bibitem[\protect\citeauthoryear{Tran and Davis}{Tran and
  Davis}{2008}]{davisMLN2008}
{\sc Tran, S.~D.} {\sc and} {\sc Davis, L.~S.} 2008.
\newblock Event modeling and recognition using markov logic networks.
\newblock In {\em Proceedings of European Conference on Computer Vision
  (ECCV)}. 610--623.

\bibitem[\protect\citeauthoryear{Vail, Veloso, and Lafferty}{Vail
  et~al\mbox{.}}{2007}]{vail2007conditional}
{\sc Vail, D.}, {\sc Veloso, M.}, {\sc and} {\sc Lafferty, J.} 2007.
\newblock Conditional random fields for activity recognition.
\newblock In {\em Proceedings of International Conference on Autonomous Agents
  and Multiagent Systems (AAMAS)}. ACM, 1--8.

\bibitem[\protect\citeauthoryear{Valiant}{Valiant}{1979}]{disjointsum}
{\sc Valiant, L.~G.} 1979.
\newblock The complexity of enumeration and reliability problems.
\newblock {\em SIAM Journal on Computing\/}~{\em 8}, 410--421.

\bibitem[\protect\citeauthoryear{Wang and Domingos}{Wang and
  Domingos}{2008}]{wang2008hybrid}
{\sc Wang, J.} {\sc and} {\sc Domingos, P.} 2008.
\newblock Hybrid {M}arkov logic networks.
\newblock In {\em Proceedings of Conference on Artificial Intelligence (AAAI)}.
  1106--1111.

\bibitem[\protect\citeauthoryear{Wu, Lian, and Hsu}{Wu
  et~al\mbox{.}}{2007}]{wu2007joint}
{\sc Wu, T.}, {\sc Lian, C.}, {\sc and} {\sc Hsu, J.} 2007.
\newblock Joint recognition of multiple concurrent activities using factorial
  conditional random fields.
\newblock In {\em Proceedings of AAAI Workshop on Plan, Activity, and Intent
  Recognition}.

\end{thebibliography}

\end{document}